\documentclass[11pt]{article}

\usepackage[final]{acl}
\usepackage{times}
\usepackage{latexsym}
\usepackage[T1]{fontenc}
\usepackage[utf8]{inputenc}
\usepackage{microtype}
\usepackage{inconsolata}
\usepackage{graphicx}
\usepackage{hyperref}       
\usepackage{url}           
\usepackage{booktabs}      
\usepackage{amsfonts}      
\usepackage{nicefrac}      
\usepackage{xcolor}        
\usepackage{amsmath,amsfonts,bm}
\usepackage{graphicx}
\usepackage{multirow}
\usepackage{makecell}
\usepackage{array}
\usepackage[para]{threeparttable}
\usepackage{booktabs}
\usepackage{tabularx}
\usepackage{xspace}
\usepackage{caption}
\usepackage{subcaption}
\usepackage{longtable}
\usepackage{hyperref}
\usepackage{bbm}
\usepackage{bm}
\usepackage{enumitem}
\usepackage{algorithm}
\usepackage{algpseudocode}
\usepackage{geometry}
\usepackage{comment}
\usepackage[colorinlistoftodos]{todonotes}
\usepackage{tablefootnote}
\usepackage{color}
\usepackage{tcolorbox}
\usepackage{tikz}
\usepackage{tabularray}
\usepackage[normalem]{ulem}
\usepackage{wrapfig}
\usepackage[T1]{fontenc}
\usepackage{listings}
\usepackage{xcolor}
\usepackage{xspace}

\newcommand{\wcontext}{\textsc{Ctx-LLM}\xspace}
\newcommand{\wocontext}{\textsc{NoCtx-LLM}\xspace}
\newcommand{\wvlm}{\textsc{Ctx-VLM}\xspace}
\newcommand{\wovlm}{\textsc{NoCtx-VLM}\xspace}

\title{Instruction Tuning with and without Context: Behavioral Shifts and Downstream Impact}

\author{
  Hyunji Lee$^{\hspace{.1em}{\boldsymbol{1}}}$ \quad
  Seunghyun Yoon$^{\hspace{.1em}\boldsymbol{2}}$ \quad
  Yunjae Won$^{\hspace{.1em}\boldsymbol{3}}$ \quad
  Hanseok Oh$^{\hspace{.1em}\boldsymbol{4}}$ \quad \\ 
  \textbf{Geewook Kim}$^{\hspace{.1em}\boldsymbol{4,5}}$ \quad  
  \textbf{Trung Bui}$^{\hspace{.1em}\boldsymbol{2}}$ \quad  
  \textbf{Franck Dernoncourt}$^{\hspace{.1em}\boldsymbol{2}}$ \quad 
  \textbf{Elias Stengel-Eskin}$^{\hspace{.1em}\boldsymbol{1}}$ \quad \\ 
  \textbf{Mohit Bansal}$^{\hspace{.1em}\boldsymbol{1}}$ \quad 
  \textbf{Minjoon Seo}$^{\hspace{.1em}\boldsymbol{3}}$ \quad \\
  \vspace{0.1em} \\
  $^{1}$ UNC Chapel Hill \hspace{1.0em} $^{2}$ Adobe Research \hspace{1.0em} $^{3}$ KAIST AI \\ 
  $^{4}$ Mila – Quebec AI Institute \hspace{1.0em}
  $^{5}$ NAVER Cloud AI \hspace{3.0em}  \\
}

\begin{document}    
\maketitle
\begin{abstract}

Instruction tuning is a widely used approach to improve the instruction-following ability of large language models (LLMs). Instruction-tuning datasets typically include a mixture of context-augmented and context-free examples, yet prior work has largely combined these data types without examining their distinct effects. In this paper, we investigate how training LLMs with or without context affects model behavior and downstream performance. First, in the text domain, we show that LLMs trained with context attend more strongly to the provided knowledge, achieving better grounding. We also observe that context-augmented training shifts how LLMs use knowledge: models store and leverage less on parametric knowledge and instead depend more on the provided context. Second, we observe that using LLM trained with context-augmented data as the backbone for vision-language models reduces hallucination and improves grounding in the visual domain. Finally, we explore practical strategies for real-world deployments where context availability varies. We show that maintaining separate context-augmented and context-free models and routing inputs between them yields more robust overall performance than training a single mixed model, as it better preserves their complementary strengths\footnote{\url{https://github.com/kaistAI/Effects-of-Context-in-Instruction-Tuning}}.

\end{abstract}

\section{Introduction}

Large language models (LLMs) are often adapted to follow user instructions through instruction tuning, which finetunes the model on pairs of instructions and responses so that the models learn to operate in desired ways~\citep{wei2021finetuned, ouyang2022training, sanh2021multitask}.
Instruction-tuning datasets~\citep{alpaca, DatabricksBlog2023DollyV2} vary in whether each example is augmented with external context (e.g., supporting documents) or is presented without any context.
In practice, researchers often combine both data types to expose models to a broad range of instructions: context-augmented examples can help disambiguate prompts, support tasks such as summarization, and teach models to use provided evidence, while context-free examples encourage open-ended reasoning and general instruction following.
Yet most prior work simply \textit{mixes} these two data sources during training without examining their individual effects.
This leaves an open question: \textit{how does training with context-augmented or context-free instruction tuning data shape model knowledge and behaviors, and how do these differences transfer to other applications such as vision–language models?}

In this paper, we analyze these questions by comparing two LLM variants, \wcontext, which is trained on context-augmented instruction data, and \wocontext, which is trained on context-free instruction data.
Specifically, we address three research questions: 
(\textbf{\textsc{RQ1}}) How do \wcontext and \wocontext differ in performance and knowledge use?
(\textbf{\textsc{RQ2}}) How does using \wcontext and \wocontext as backbones for vision-language adaptation affect performance on vision-language tasks?
(\textbf{\textsc{RQ3}}) How can these insights guide when to use \wcontext and \wocontext, and how to combine them effectively for downstream applications?  

We observe that \wcontext achieves higher performance on information-seeking tasks when context is provided compared to \wocontext, as it implicitly learns to attend more strongly to the given evidence.
Also, \wcontext maintains strong general language understanding, demonstrating solid language comprehension and reasoning ability.
However, \wcontext performance drops on knowledge-intensive tasks without context, where the model must rely on its own parametric knowledge.
Our analysis suggests that this degradation arises because training with context shifts the model’s reliance away from internal parametric memory and toward the provided context.

Next, we analyze how differences in LLM behavior, driven by the characteristics of the instruction-tuning data, affect performance on vision-language tasks.
We compare models that use either \wcontext or \wocontext as the \textit{backbone} for vision-language adaptation, applying the same vision-language alignment procedure in both cases.
We find that using \wcontext as the backbone (\wvlm) reduces hallucination and improves grounding in the input image, showing that the grounding ability learned in the text domain successfully transfers to the visual domain as well. 
Moreover, \wvlm maintains stable factual accuracy across long generated responses, including facts expressed later in the sequences, where models often degrade and hallucination~\citep{lee2024toward}.
Finally, \wvlm preserves strong performance on general vision-language understanding and reasoning benchmarks.

Finally, we investigate how these insights can inform the design and adaptation of instruction-tuning datasets for different downstream applications.
Our earlier analysis shows that \wcontext is better suited for tasks that use provided context as the knowledge source, while \wocontext excels when tasks must rely on the model’s internal knowledge.
However, many real-world applications require \textit{both} capabilities.
We therefore examine two approaches.
First, we study the common practice of \textit{mixing} context-augmented and context-free data into a single training set. 
Varying their ratio with a fixed total size shows that increasing context-augmented data improves performance on context-based tasks but slightly degrades parametric knowledge use, with roughly a 50/50 mix giving the most balanced single-model performance.
Second, we investigate a \textit{routing} setup: we train \wcontext and \wocontext separately and route inputs to the appropriate model using a simple heuristic of whether external context is provided. 
The routing setup consistently outperforms the mixed model, suggesting that keeping the two models separate and routing inputs is a practical way to preserve both context-based and parametric knowledge use.

\section{Experimental Setup} \label{sec: setup}

In Section~\ref{expsec: comparison}, we compare models trained on instruction tuning dataset with and without context.
The following sections describe the experimental setup used to train and evaluate these two LLMs in both the text-only domain~(Section~\ref{expsec: text}) and the vision-language domain~(Section~\ref{expsec: vlm}). See Appendix~\ref{app_sec: exp_setup} for more details. 

\subsection{Comparison Models: \wcontext vs. \wocontext}
\label{expsec: comparison}
Instruction tuning a model without context~(\wocontext) involves training it to generate responses $r$ given an instruction $i$ by minimizing the negative log-likelihood of the response tokens:
\[
L(\theta) \approx -\mathbb{E}_{(i, r)} \sum_{t_k\in r} \log P_{\theta}(t_k \mid i, t_{<k}),
\]
where $P_{\theta}(t_k \mid i, t_{<k})$ is the probability assigned by an autoregressive language model with parameters $\theta$ to the next token $t_k$, conditioned on the instruction $i$ and the preceding tokens $t_{<k}$.

Instruction tuning a model with context~(\wcontext) follows the same objective but includes an additional input $c$, representing external knowledge. During training, the context is provided alongside the instruction and prepended to the target response:
\[
L(\theta) \approx - \mathbb{E}_{(i, c, r)} \sum_{t_k \in r} \log P_{\theta}(t_k \mid i, c, t_{<k}).
\]

Following the design of previous work~\citep{asai2024selfrag, lee2024semiparametric} and our dataset construction, when training \wcontext, the relevant context $c$ is prepended to the corresponding sentences in the target response $r$. 
The loss is computed only on the response tokens, exclduing context itself. 
We provide additional experiments validating this design choice in Appendix~\ref{app_sec: exp_setup}.

At inference time, both models follow the same generation procedure. If external knowledge is available, it is provided as context; otherwise, the model generates a response from the instruction alone.

\subsection{Text Domain} \label{expsec: text}

\paragraph{Datasets \& Evaluation Metrics}
For training, we use the 29k dataset from Self-RAG~\citep{asai2024selfrag}, constructed by augmenting instruction tuning datasets with relevant context identified at the sentence level and incorporated when available\footnote{See Appendix~\ref{app_sec: exp_setup} for details on filtering over Self-RAG training datasets.}.
For evaluation, we experiment over 11 information-seeking datasets, including NQ~\citep{kwiatkowski-etal-2019-natural}, TriviaQA~(TQA)~\citep{2017arXivtriviaqa}, zsRE~\citep{levy2017zero}, T-rex~\citep{elsahar-etal-2018-rex}, and HotpotQA~(HQA)~\citep{yang2018hotpotqa}, using the versions provided in KILT~\citep{petroni-etal-2021-kilt}. 
For all experiments, we use the context provided in the original dataset.
We also include DROP~\citep{dua2019drop}, SQuAD~\citep{rajpurkar2016squad}, SWDE~\citep{lockard-etal-2019-openceres}, and FDA~\citep{arora2023language}, for which we use the version curated by Based~\citep{arora2024just}. 
Additionally, we evaluate on two benchmarks specifically designed to highlight grounding failures in language models: NQ Conflict (NQ-C)~\citep{zhou2023context} and the dataset from CORG~\citep{corg}. 
For the dataset from CORG, we follow \cite{corg} in reporting the D-F1 metric, and for the rest, we evaluate using answer accuracy.
We evaluate over 7 downstream tasks for general LLM performance: PIQA~\citep{Bisk2020}, Social IQa~\citep{sap2019social}, Winogrande~\citep{sakaguchi2019winogrande}, HellaSwag~\citep{zellers2019hellaswag}, LAMBADA-OpenAI~\citep{lambada}, ARC-Challenge, and ARC-Easy~\citep{Clark2018ThinkYH}) using lm-evaluation-harness~\citep{eval-harness}.

\paragraph{Training Details}
We conduct experiments on three pretrained models: Llama 2 7B~\citep{touvron2023llama}, Llama 3.1 8B~\citep{grattafiori2024llama}, and Qwen 2.5 7B~\citep{yang2024qwen2}\footnote{If not specified, we use Llama 3.1 8B as the base model for all analyses, with full-parameter fine-tuning.}. 
We train the full model for three epochs with a batch size of 128, a learning rate of 2e-5, and the AdamW optimizer. All training are conducted using 4 NVIDIA A100 80G GPUs. 

\subsection{Vision-Language Domain}  \label{expsec: vlm}
\paragraph{Datasets \& Evaluation Metrics}
When training the vision-language alignment, we use the training dataset from LLaVA~\citep{liu2023llava} for both the pretraining and finetuning stages.
We evaluate over four hallucination benchmarks AMBER~\citep{wang2023llm}, POPE~\citep{li2023evaluating}, ImageInWords~\citep{garg2024imageinwords}, and LLaVA-Wild~\citep{liu2023llava} to measure its grounding ability to provided image.
For POPE, we report the average F1 score across all splits.
For ImageInWords, we adopt the evaluation from CapMAS~\citep{lee2024toward}, a GPT-based approach for fine-grained factuality assessment.
For LLaVA-Wild, to better target hallucination detection, we modify the rubric to explicitly penalize hallucinations.
We evaluate over four downstream tasks for general VLM performance: MMBench~\citep{MMBench}, ScienceQA~\citep{lu2022learn}, MME~\citep{sun2023mme}, and GQA~\citep{hudson2019gqa}.

\paragraph{Training Details}

For all models, we follow a widely used two-stage process~\citep{chen2024far, liu2023llava}: first, a pretraining stage for feature alignment, and second, an end-to-end fine-tuning stage. 
Specifically, we follow the training configurations introduced in LLaVA~\citep{liu2023llava}, and conduct all experiments on 8 NVIDIA A100 80G GPUs.
Note that the vision-language alignment procedure is identical across all experiments, and only the backbone for the LLM varies.

\begin{table*}[t!]
\centering
\fontsize{7}{10} \selectfont
    \begin{tabular}{cc|ccccc|cc|cccc|c}
    \toprule
    BaseModel & ModelType & NQ & TQA & zsRE & T-rex & HQA &  NQ-C & Corg  & Drop & Squad & SWDE & FDA & \textbf{\textit{Avg}} \\ 
    \midrule
    \multirow{2}{*}{Llama2} 
    & \wocontext & 42.3 & 69.0 & 51.2 & \textbf{69.8} & 45.5 &  54.1 &17.3 & 33.8 & 42.9  & \textbf{82.3} & 73.4 & 52.9\\
    & \wcontext & \textbf{55.8} & \textbf{72.1} & \textbf{65.1} & \text{60.2} & \textbf{49.0} & \textbf{75.1} & \textbf{19.1} & \textbf{38.7} & \textbf{58.3} & 81.0 & \textbf{76.3} & \textbf{59.2} \\ 
    \midrule
    \multirow{2}{*}{Llama3.1} 
    &  \wocontext & \textbf{48.0} & 70.2& 57.5& 62.7 & 48.0 & 60.7 & 18.0 & 39.0 & 64.9 & 92.0 & 74.0 & 57.7 \\ 
    & \wcontext & \text{46.2} & \textbf{72.6} & \textbf{62.6} & \textbf{63.4} & \textbf{50.6} & \textbf{72.3} & \textbf{19.4} & \textbf{44.0} & \textbf{69.5} & \textbf{95.0} & \textbf{80.9} & \textbf{61.5}\\ 
    \midrule
    \multirow{2}{*}{Qwen2.5} 
    &  \wocontext & 57.2 & 73.5 & 67.0 & 68.9 & \textbf{60.0} & 58.2 &17.9 & 30.9 & \textbf{62.7} &84.1& 81.9 & 60.2\\ 
    &  \wcontext & \textbf{66.1} & \textbf{84.6} & \textbf{72.8} & \textbf{74.6} & \text{54.2} &  \textbf{71.0} & \textbf{20.7} & \textbf{49.5} &\text{60.5} &\textbf{88.2}&\textbf{89.4} & \textbf{66.5}\\ 
    \bottomrule
    \end{tabular}
\caption
     {
     Performance comparison of instruction-tuned models trained with context (\wcontext) vs. without context (\wocontext) across 11 information-seeking datasets, using three base models: Llama2 7B, Llama3.1 8B, and Qwen2.5 7B.
     } 
\label{table: main_table}
\end{table*}

\section{RQ1: How do \wcontext and \wocontext differ in performance and knowledge use?}

In this section, we investigate how training an LLM on instruction-tuning datasets with context~(\wcontext) versus without context~(\wocontext) affects its behavior on information-seeking tasks and general language understanding.

\begin{figure}[t!]
    \centering    
    \includegraphics[width=\linewidth]{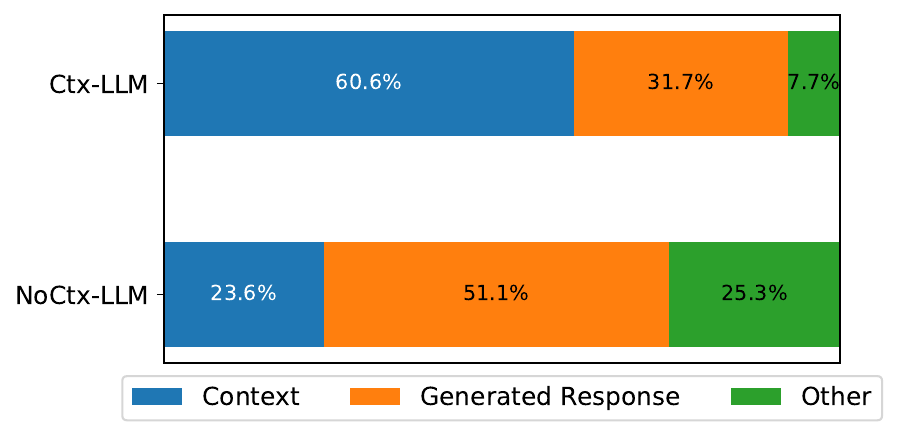}
    \caption{
    Avg. rate at which the model attends most to each input segment: context, generated response, or other (e.g., system prompts) during generation.
    } 
    \label{fig: attn_rate}
\end{figure}

\subsection{\wcontext Improves Grounding}
Table~\ref{table: main_table} shows that \wcontext, LLMs trained with context, consistently achieve higher performance on information-seeking datasets when provided with relevant context at inference time. 
Across a range of pretrained backbones (Llama 2 7B, Llama 3.1 8B, and Qwen 2.5 7B), \wcontext yields an average absolute improvement of 5.5\% over \wocontext, which are trained without context.
The gains are especially pronounced on benchmarks such as NQ-C and CORG, which require handling counterfactual or complex knowledge in the provided context; \wcontext achieves an average of 8.6\% absolute improvement compared to an average of 4.8\% improvement on the remaining datasets. 
These results suggest that training models on instruction-tuning datasets containing context strengthens their grounding ability, even in cases where standard models often misinterpret or generate incorrect responses.

\subsection{Shift in Generation Behavior Induced by Training on Context-Augmented Data} \label{behavior: attn}

We observe that while training with context does not \textit{explicitly} guide the model to attend more strongly to the external context, the inclusion of relevant context in training instances implicitly encourages this behavior and enhances grounding; \wcontext shows better grounding ability than \wocontext.
Figure~\ref{fig: attn_rate} shows the average ratio of which part of the input (Context, Generated Response, or Other) the models attend to the most when generating responses for instances in NQ. 
For each generated token, we identify the input segment that receives the highest attention weight, compute the proportion of tokens attending to each segment per instance, and then average those proportions across all instances.

Results show that \wcontext assigns greater attention to the given context (\textit{blue}), whereas \wocontext tends to focus more on its own previously generated responses (\textit{orange}).
A similar pattern appears in the full attention map provided in Appendix~\ref{app_sec: behavior}.
These results suggest that training a model with context included induces a mechanistic shift in the model's generation process, leading the model to assign greater attention to relevant context tokens rather than to self-generated content, thereby producing more grounded outputs.

\begin{figure}[t!]
    \centering
\includegraphics[width=0.45\textwidth]{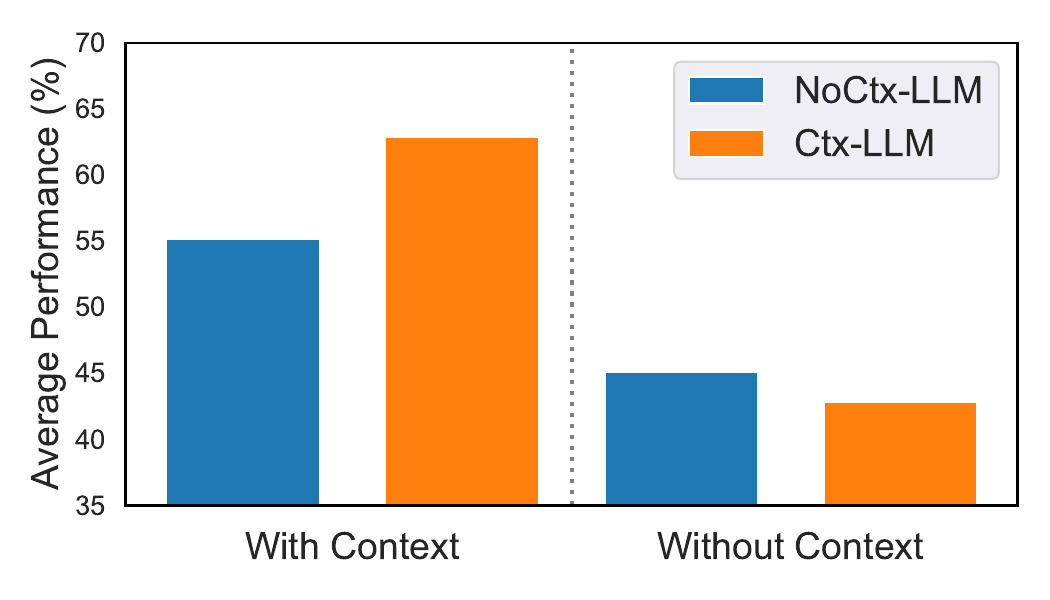}
\caption{Average Performance of \wcontext and \wocontext across inference setups. The x-axis indicates whether context is provided at inference: \textit{With Context} uses external context, while \textit{Without Context} requires the model to rely on its own parametric knowledge.} 
\label{fig:with_without_context}
\vspace{-1em}
\end{figure}

\begin{figure}[t]
\centering
\includegraphics[width=\linewidth]{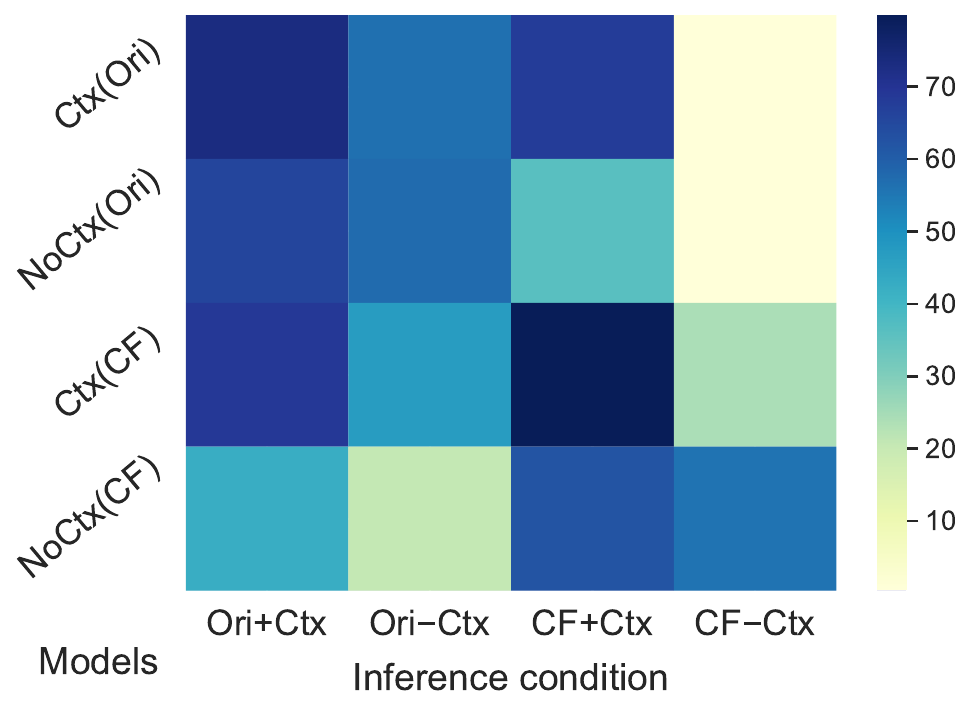}
\caption{
Accuracy by models trained on different datasets (\textit{Models}) and evaluated under different inference conditions (\textit{Inference condition}). 
\textit{Original (Ori)} refers to knowledge aligned with the model's parametric knowledge, while \textit{Counterfactual (CF)} denotes counterfactual knowledge. 
``+'' or Ctx indicates that context is provided; ``-'' or NoCtx indicates no context is provided during training or inference.
}
\label{fig:heatmap}
\vspace{-1em}
\end{figure}

\subsection{Analysis over Impact of Context-Augmented Training on the Use of Parametric Knowledge}
\label{subsec: analysis_behavior}

When evaluating on datasets where no relevant context is provided, which forces models to rely solely on their parametric knowledge, we observe that \wcontext performs less effectively than \wocontext (Figure~\ref{fig:with_without_context}). 
We hypothesize that training with context-augmented data shifts how models use knowledge: \wocontext tends to encode more information directly in its parameters, while \wcontext learns to depend on provided context. This aligns with our earlier observation that \wcontext attends more strongly to external evidence.

To test this hypothesis, we manipulate the relationship between the knowledge contained in the training data and the model’s parametric knowledge. 
In addition to the \textit{original (Ori)} dataset, where the provided context and gold answer align with the model’s prior knowledge, we construct a \textit{counterfactual (CF)} dataset, where the context and answer contradict the model’s parametric knowledge. 
We then train four models by crossing context availability (\textsc{Ctx} vs. \textsc{NoCtx}) with knowledge alignment (Ori vs. CF), and evaluate each model under four inference settings that vary (i) whether external context is available (+Ctx / –Ctx) and (ii) whether the context and answer are original or counterfactual (Ori / CF). 

Results in Figure~\ref{fig:heatmap}) suggest that training on context-augmented (\textsc{Ctx}) versus context-free data (\textsc{NoCtx}) leads to \textit{differences in how models internalize and use knowledge}. 
\textsc{NoCtx} models tend to memorize training knowledge more strongly, achieving the highest score in inference setting where they must rely on their parametric knowledge of that same knowledge.
Specifically, when trained on counterfactual knowledge, \textsc{NoCtx} models show greater forgetting of prior knowledge and stronger memorization of the updated training signal, whereas \textsc{Ctx} models retain more of their original knowledge but memorize counterfactual knowledge less strongly. 
In addition, \textsc{Ctx} models consistently achieve higher performance whenever context is available, regardless of whether the context supports original or counterfactual answers.

These findings suggest that context-augmented training shifts the model’s reliance from parametric memory toward external context. 
Models trained with context become better at using retrieved evidence but may encode less new knowledge directly in their parameters compared to models trained without context.
Details of counterfactual dataset construction and results are in Appendix~\ref{app_behavior: counterfactual_dataset}.

\begin{table*}[t!]
\centering

\fontsize{7}{10} \selectfont
    \begin{tabular}{c|c|c|cccccc|c|c}
    \toprule
    \multirow{2}{*}{BaseModel}  & \multirow{2}{*}{ModelType} & Pope & \multicolumn{6}{c}{Amber} & Llava-W & Caption\\
    \cmidrule{3-11}
     & & F1 & CHAIR~($\downarrow$) & Cover~($\uparrow$) & Hal~($\uparrow$) & Cog~($\downarrow$) & F1~($\uparrow$) & AMBER~($\uparrow$) & GPT4-Eval & F1\\
    \midrule
    \multirow{2}{*}{Llama2} 
    & \wovlm & 84.7 & 9.3 & 47.8 & \textbf{38.8} & 5.0 &  65.0 & 77.9 & 53.4 &  54.1\\
    & \wvlm & \textbf{85.5} & \textbf{7.3} & \textbf{48.0} & 30.5 & \textbf{3.4} & \textbf{71.4} & \textbf{82.1} & \textbf{70.9}& \textbf{55.9}\\
    \midrule
    \multirow{2}{*}{Llama3.1} 
    & \wovlm & 87.3 & 9.1& 53.8&39.8  &5.9 & 66.5 & 78.7 & 55.7 & 54.5 \\
    & \wvlm & \textbf{87.7} & \textbf{8.6} & \textbf{54.6} & \textbf{47.2} & \textbf{4.9} & \textbf{71.7} & \textbf{81.6} & \textbf{74.2} & \textbf{56.8} \\
    \midrule
    \multirow{2}{*}{Qwen2.5} 
    & \wovlm & 87.9 & 8.5 & 49.1& 34.4& \textbf{4.8} & 70.8 & 81.2 & 60.1 & 56.7 \\
    & \wvlm & \textbf{88.8} & \textbf{7.2} & \textbf{52.9} & \textbf{40.2} & 5.1 & \textbf{72.3} & \textbf{82.6} & \textbf{78.5} & \textbf{58.7} \\
    \bottomrule
    \end{tabular}
\caption
     {
     Performance of VLM using \wocontext as the LLM backbone (\wovlm) and \wcontext as the LLM backbone (\wvlm) across four hallucination benchmarks, using three base models (Llama2 7B, Llama3.1 8B, Qwen 2.5 7B). The first row indicates the evaluation dataset, and the second row shows the metric.
     } 
\label{table: vlm_hallucination}
\end{table*}

\begin{table}[t!]
\centering

\fontsize{7}{10} \selectfont
    \begin{tabular}{c|ccccccc|c}
    \toprule
    Ctx & PI & SI & WI & HS& LA & AC& AE & \textbf{\textit{Avg}}\\ 
    \midrule
    \textsc{F} & 81.9 & 48.6 & 72.6 & \textbf{80.6} & 75.6 & \textbf{56.7} & 82.2 & 68.2 \\ 
    \textsc{T}& \textbf{82.9} & \textbf{49.0} & \textbf{73.4} & 80.3 & \textbf{76.3} & 56.1 & \textbf{83.0} & \textbf{68.4} \\
    \bottomrule
    \end{tabular}
\caption
     {Performance of \wcontext (Ctx=T) and \wocontext (Ctx=F) using Llama3.1 8B as base model, across seven widely used downstream benchmarks. PI is PiQA, SI is SocialIQA, WI is Winogrande, HS is Hellaswag, LA is LAMBADA-OpenAI, AC is ARC-challenge, and AE is ARC-Easy.} 
\label{table: standard_downstream}
\end{table}
\subsection{\wcontext show high performance in general language understanding.} \label{subsec: text_general}
To compare model behavior or performance of \wcontext and \wocontext over other tasks apart from information-seeking tasks, we evaluate them on seven widely used downstream benchmarks: PIQA, Social IQa, Winogrande, HellaSwag, LAMBADA-OpenAI, ARC-Challenge, and ARC-Easy.
Results in Table~\ref{table: standard_downstream} show that \wcontext tend to show higher or comparable performance compared to \wocontext, achieving an average score of 68.4 compared to 68.2. 
This suggests that training with context-augmented data tends to preserve or even enhance general language understanding across diverse tasks. 

\section{RQ2: How does using \wcontext or \wocontext as the backbone for vision-language adaptation influence performance on vision-language tasks?}

In this section, we analyze how training an LLM with or without context (\wcontext vs. \wocontext) affects its performance when used as the backbone for vision-language adaptation.
We compare two configurations: \wvlm, which uses an LLM trained with context (\wcontext), and \wovlm, which uses one trained without context (\wocontext), while keeping all other training and alignment procedures identical.

\begin{figure}[t!]
    \centering
\includegraphics[width=0.45\textwidth]{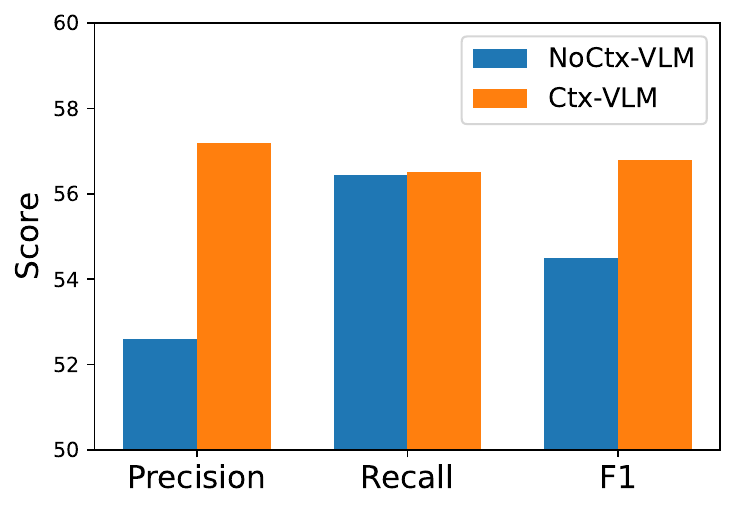}
\caption{Precision, Recall, and F1 score on the ImageInWords fine-grained captioning task, evaluated with CapMAS, comparing \wvlm and \wovlm using Llama3.1 8B as base model.} 
\label{fig: caption_perf}
\vspace{-1em}
\end{figure}

\subsection{Using \wcontext as a backbone improves grounding in vision–language models.} \label{sec: vlm}
To assess grounding in vision-language models, we evaluate \wvlm and \wovlm on four hallucination benchmarks: POPE, AMBER, LLaVA-Wild, and ImageInWords.
Table~\ref{table: vlm_hallucination} shows that \wvlm consistently outperforms \wovlm, achieving higher accuracy and reduced hallucination. 
This suggests that \wvlm generates responses that are more faithfully grounded in the visual input rather than relying primarily on its parametric knowledge.

Notably, in LLaVA-Wild, the performance gap widens when the evaluation rubric penalizes hallucination more heavily: \wvlm surpasses \wovlm by +18.1\% under the stricter rubric, compared to +8.8\% under the original rubric.
Similarly, as shown in Figure~\ref{fig: caption_perf}, \wvlm achieves stronger results on ImageInWords, with a notable gain in precision. 
These gains indicate that captions generated by \wvlm are more accurate and better grounded in the provided visual input.
Together, these results highlight the advantage of using a context-augmented backbone (\wcontext) for reducing hallucination and improving grounding in vision-language models.

We hypothesize that this improved grounding ability stems from the generalization of grounding behaviors learned during instruction tuning with context-augmented data.
Specifically, in \wcontext, models learn to effectively leverage provided external knowledge, and this grounding behavior appears to \textit{transfer when the external input shifts from textual context to visual information} in the vision-language setting (\wvlm). 

\begin{table}[t!]
\centering

\fontsize{7}{10} \selectfont
    \begin{tabular}{c|ccccc}
    \toprule
    Model & MMBench & ScienceQA & MME &GQA \\
    \midrule
    \wovlm &68.6 &78.4 & 1526.1 & 63.4\\
    \wvlm & \textbf{70.2} & \textbf{79.1}  &  \textbf{1534.4} & \textbf{64.1} \\
    \bottomrule
    \end{tabular}
    \caption
     {Comparison of \wovlm and \wvlm over four vision-language downstream tasks using Llama 3.1-8B as base model.} 
     \label{table: vlm_tasks}
\end{table}

\subsection{\wvlm show robust performance across varying response lengths.}

\begin{figure}[t!]
    \centering
\includegraphics[width=0.45\textwidth]{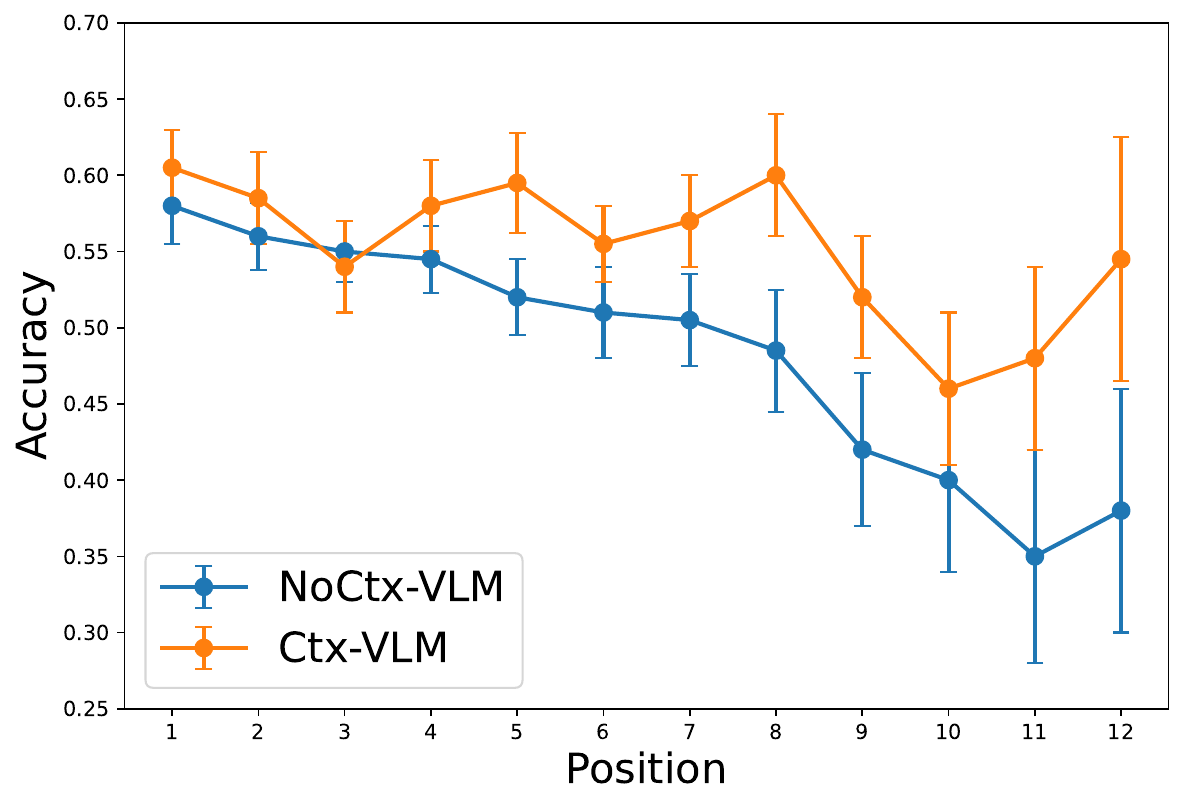}
\caption{Avg. accuracy (y-axis) of atomic facts from generated responses as a function of their position (x-axis). Error bars indicate variance.} 
\label{fig:correlation}
\vspace{-0.8em}
\end{figure}
Figure~\ref{fig:correlation} presents fine-grained captioning performance on the ImageInWords benchmark, showing accuracy as a function of the position of knowledge within the generated response, evaluated using the CapMAS method~\citep{lee2024toward}, which decomposes generated sentences into atomic fact units using GPT-4o and assesses their truthfulness based on the corresponding image and reference caption.
The x-axis in the figure indicates the position at which a fact appears in the generated caption.

\wvlm maintains more stable accuracy across different positions, especially outperforming \wovlm at later positions, where the performance gap becomes increasingly pronounced.
This suggests that \wvlm are more robust at preserving factual consistency throughout the entire generated response. 
A qualitative example of fine-grained caption in Appendix~\ref{app_sec: response_len} also illustrates this behavior; while the initial parts of the responses from \wovlm and \wvlm tend to be similar, their outputs diverge notably toward the end where \wovlm generates an incorrect response and \wvlm provides a more detailed observation.

\subsection{\wvlm shows robust performance on general VLM downstream tasks}

To evaluate the performance of \wvlm and \wovlm beyond hallucination benchmarks, we test both models on four representative vision-language benchmarks: MMBench, ScienceQA, MME, and GQA.
Table~\ref{table: vlm_tasks} shows that \wvlm achieves performance comparable to or higher than \wovlm, indicating that its improved grounding ability and reduced hallucination do not come at the expense of general vision-language understanding and reasoning.

\begin{figure}[t!]
\begin{center}
\includegraphics[width=0.45\textwidth]{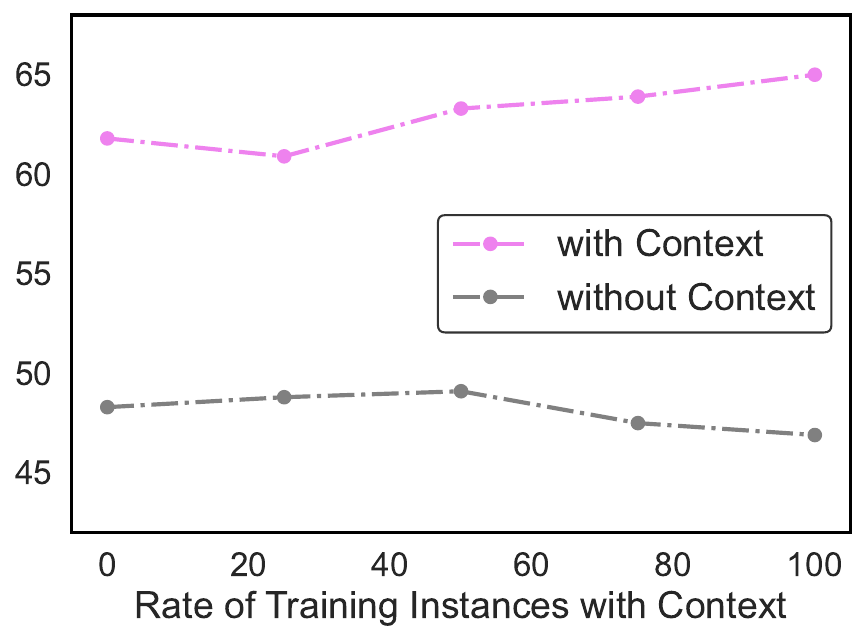}
\end{center}
\caption{
Performance of models (y-axis) trained with different proportions of context-augmented training instances (x-axis). The pink line shows performance when inference is performed with relevant context, while the gray line shows performance when inference is performed without context.
}
\label{fig: instance_ratio}
\vspace{-1em}
\end{figure}

\section{RQ3: How can these insights guide when to use each model and combine them effectively for downstream applications?}

Our earlier experiments show that \wcontext demonstrates strong grounding ability\footnote{In Appendix~\ref{app_sub: inference_time_grounding}, we further observe that using \wcontext with inference-time grounding methods yields higher performance than using \wocontext, reinforcing that \wcontext is better suited for grounding-based applications.} while maintaining general language understanding, making it a suitable choice for applications where relevant external context is available at inference time (e.g., vision–language tasks, retrieval-augmented LLMs). 
Conversely, \wocontext achieve higher performance when they must rely on their own parametric knowledge without access to external evidence. 
However, many real-world scenarios require a model to effectively use \textit{both} provided context and its internal knowledge.

In this section, we examine two approaches for combining the strengths of \wcontext and \wocontext in such applications.
The first is \textit{mixture training}, a common practice in instruction tuning, where a single model is trained on a mix of context-augmented and context-free examples.  
The second is \textit{routing-based inference}, where two models, \wcontext and \wocontext, are trained separately, and inputs are routed to the appropriate model using a simple heuristic of whether relevant context is available.

\paragraph{Training with Mixture of Context-augmented and Context-free Data}

We investigate how the proportion of context-augmented versus context-free training examples affects a model's ability to ground in external context while retaining parametric knowledge.
We keep the total number of training examples fixed and vary the proportion that is context-augmented: 0\%, 25\%, 50\%, 75\%, and 100\%.
As shown in Figure~\ref{fig: instance_ratio}, increasing the rate of context-augmented training data generally improves performance when context is available at inference, while causing slight degradation when no context is provided. 
Notably, a 50\% mix achieves the most balanced performance, maintaining strong grounding ability when context is available while preserving performance on knowledge-intensive tasks without context.

\paragraph{Routing Inputs Based on Context Availability}

We experiment with a routing setup where inputs are directed to either \wcontext or \wocontext depending on whether relevant context is provided in inference step.
Results in Figure~\ref{fig: practical} show that routing (purple line) performs well in both when inference with and without context. 
This suggests that the two models could be used in complementary ways; for example, when designing or extending mixture-of-experts architectures, one could include \wcontext and \wocontext as separate experts to leverage both grounding ability and parametric knowledge. 
We observe that this approach (56.7) consistently outperforms any single mixed model (55.6), suggesting that in practical deployments it is often more effective to \textit{maintain separate experts and route between them} rather than rely on a single model trained on a mixed dataset. Performance scores are averaged over nine evaluation datasets\footnote{We exclude NQ-C and Corg from the 11 information-seeking datasets, as counterfactual answers are not reliably answerable without context.} covering both context-available and context-free inference settings.

We also tested a LoRA-based approach, where everything is equal except that it is trained with LoRA rather than full parameters to make it lightweight. We observe similar trend with when training full parameters, suggesting a practical alternative to fully maintaining two models. 
More details regarding results are in Appendix~\ref{app_sub: lora}.
Future work could extend to using a trainable router to decide dynamically whether context would improve performance, rather than relying on the simple heuristic of context presence.

\section{Related Works}
\paragraph{Instruction Tuning and Its Impact on Context Awareness and Knowledge Use}

Several works have shown that instruction tuning influences on how LLMs use their parametric knowledge and given input context. 
Recent studies show loss of context awareness after instruction tuning\footnote{Please note that the training dataset used here do not contain external knowledge (\wocontext setting)}.
\citet{goyal2024context} find that models tend to be less reliant on provided context under knowledge conflict as instruction tuning progresses.
Similarly, \citet{wang2024loss} attributes this loss to role biases introduced by chat-style prompting templates.
Other analysis explore how instruction tuning reshapes behavioral and representational properties of LLMS in attention distribution or attribution~\citep{wu2023language, gao2023roles}.
Another line of work aim to enhance the model's external context utilization through instruction tuning: overcoming the lost-in-the-middle problem~\citep{liu2023lost} in long-context inputs~\citep{an2024make, begin2025pause, he2023never} or better grounding on given context~\citep{lee2024semiparametric, asai2024selfrag, Tian2023FinetuningLM, luo2023sail}.
Together, these studies examine how instruction tuning affects the utilization of parametric knowledge versus user-provided context.
Our work extends this understanding by examining new axes of how training on datasets with or without relevant context affects a model's grounding ability and its use of parametric knowledge.

\paragraph{Improving Grounding in Language and Vision–Language Models}
Prior work have explored multiple directions to strengthen grounding.
Inference-time methods modify decoding or introduce post-hoc revision pipelines to better incorporate external knowledge~\citep{shi2023trusting, wang2024adacad, gao2022rarr, chern2023factool}.
Training-time approaches aim to align models to external evidence, for example using preference optimization for factuality~\citep{Tian2023FinetuningLM} or integrating retrieval and self-critique signals as in Self-RAG~\citep{asai2024selfrag}.

Improving grounding ability is also crucial for vision-language models (VLMs) to ensure that generated responses are based on the image rather than relying on the language model to produce plausible outputs, but not based on the image. 
Prior work has explored several directions for enhancing grounding in VLMs, including adjusting decoding~\citep{Favero2024MultiModalHC, Leng2023MitigatingOH}, curating high-quality datasets~\citep{liu2023aligning, Li2023M3ITAL}, and developing training strategies to strengthen visual grounding~\citep{Sun2023AligningLM, ouali2024clip}.

\section{Conclusion}

In this paper, we studied the impact of training LLMs with context-augmented data (\wcontext) versus context-free data (\wocontext).
We observed that training with context shifts how model uses knowledge: reducing reliance on its parametric knowledge and encouraging stronger use of the provided context. 
This behavior generalizes beyond the text domain to the visual domain, leading to improved performance on hallucination benchmarks. 
Moreover, our exploration on practical deployment strategies suggests that rather than mixing both data types into a single model, maintaining separate \wcontext and \wocontext and routing yields stronger overall performance.

\newpage
\section*{Limitations}

Our experiments are conducted on 7B scale models due to computational constraints.
However, we tested over three different base models and observed a consistent trend, suggesting that our findings are likely to generalize to various, larger models.
Additionally, our study assumes that the provided context is reliable and relevant. Future work could explore scenarios where the context may be unreliable, noisy, or partially irrelevant, requiring the model to assess the trustworthiness of external information before grounding on it.

\bibliography{custom}

@article{hu2022lora,
  title={Lora: Low-rank adaptation of large language models.},
  author={Hu, Edward J and Shen, Yelong and Wallis, Phillip and Allen-Zhu, Zeyuan and Li, Yuanzhi and Wang, Shean and Wang, Lu and Chen, Weizhu and others},
  journal={ICLR},
  volume={1},
  number={2},
  pages={3},
  year={2022}
}

@inproceedings{ouali2024clip,
  title={Clip-dpo: Vision-language models as a source of preference for fixing hallucinations in lvlms},
  author={Ouali, Yassine and Bulat, Adrian and Martinez, Brais and Tzimiropoulos, Georgios},
  booktitle={European Conference on Computer Vision},
  pages={395--413},
  year={2024},
  organization={Springer}
}

@article{wang2023llm,
  title={An llm-free multi-dimensional benchmark for mllms hallucination evaluation},
  author={Wang, Junyang and Wang, Yuhang and Xu, Guohai and Zhang, Jing and Gu, Yukai and Jia, Haitao and Yan, Ming and Zhang, Ji and Sang, Jitao},
  journal={arXiv preprint arXiv:2311.07397},
  year={2023}
}

@article{lee2024semiparametric,
  title={Semiparametric Token-Sequence Co-Supervision},
  author={Lee, Hyunji and Kim, Doyoung and Jun, Jihoon and Joo, Sejune and Jang, Joel and On, Kyoung-Woon and Seo, Minjoon},
  journal={ACL},
  year={2024}
}

@article{li2023evaluating,
  title={Evaluating object hallucination in large vision-language models},
  author={Li, Yifan and Du, Yifan and Zhou, Kun and Wang, Jinpeng and Zhao, Wayne Xin and Wen, Ji-Rong},
  journal={arXiv preprint arXiv:2305.10355},
  year={2023}
}

@article{wang2024adacad,
  title={Adacad: Adaptively decoding to balance conflicts between contextual and parametric knowledge},
  author={Wang, Han and Prasad, Archiki and Stengel-Eskin, Elias and Bansal, Mohit},
  journal={arXiv preprint arXiv:2409.07394},
  year={2024}
}

@article{shi2023trusting,
  title={Trusting your evidence: Hallucinate less with context-aware decoding},
  author={Shi, Weijia and Han, Xiaochuang and Lewis, Mike and Tsvetkov, Yulia and Zettlemoyer, Luke and Yih, Scott Wen-tau},
  journal={arXiv preprint arXiv:2305.14739},
  year={2023}
}

@article{chern2023factool,
  title={FacTool: Factuality Detection in Generative AI--A Tool Augmented Framework for Multi-Task and Multi-Domain Scenarios},
  author={Chern, I and Chern, Steffi and Chen, Shiqi and Yuan, Weizhe and Feng, Kehua and Zhou, Chunting and He, Junxian and Neubig, Graham and Liu, Pengfei and others},
  journal={arXiv preprint arXiv:2307.13528},
  year={2023}
}

@article{gao2022rarr,
  title={Rarr: Researching and revising what language models say, using language models},
  author={Gao, Luyu and Dai, Zhuyun and Pasupat, Panupong and Chen, Anthony and Chaganty, Arun Tejasvi and Fan, Yicheng and Zhao, Vincent Y and Lao, Ni and Lee, Hongrae and Juan, Da-Cheng and others},
  journal={arXiv preprint arXiv:2210.08726},
  year={2022}
}

@article{Tian2023FinetuningLM,
  title={Fine-tuning Language Models for Factuality},
  author={Katherine Tian and Eric Mitchell and Huaxiu Yao and Christopher D. Manning and Chelsea Finn},
  journal={ArXiv},
  year={2023},
  volume={abs/2311.08401},
  url={https://api.semanticscholar.org/CorpusID:265158181}
}

@article{Sun2023AligningLM,
  title={Aligning Large Multimodal Models with Factually Augmented RLHF},
  author={Zhiqing Sun and Sheng Shen and Shengcao Cao and Haotian Liu and Chunyuan Li and Yikang Shen and Chuang Gan and Liangyan Gui and Yu-Xiong Wang and Yiming Yang and Kurt Keutzer and Trevor Darrell},
  journal={ArXiv},
  year={2023},
  volume={abs/2309.14525},
  url={https://api.semanticscholar.org/CorpusID:262824780}
}

@article{Leng2023MitigatingOH,
  title={Mitigating Object Hallucinations in Large Vision-Language Models through Visual Contrastive Decoding},
  author={Sicong Leng and Hang Zhang and Guanzheng Chen and Xin Li and Shijian Lu and Chunyan Miao and Li Bing},
  journal={2024 IEEE/CVF Conference on Computer Vision and Pattern Recognition (CVPR)},
  year={2023},
  pages={13872-13882},
  url={https://api.semanticscholar.org/CorpusID:265466833}
}

@article{Li2023M3ITAL,
  title={M3IT: A Large-Scale Dataset towards Multi-Modal Multilingual Instruction Tuning},
  author={Lei Li and Yuwei Yin and Shicheng Li and Liang Chen and Peiyi Wang and Shuhuai Ren and Mukai Li and Yazheng Yang and Jingjing Xu and Xu Sun and Lingpeng Kong and Qi Liu},
  journal={ArXiv},
  year={2023},
  volume={abs/2306.04387},
  url={https://api.semanticscholar.org/CorpusID:259095896}
}

@article{liu2023aligning,
  title={Aligning Large Multi-Modal Model with Robust Instruction Tuning},
  author={Liu, Fuxiao and Lin, Kevin and Li, Linjie and Wang, Jianfeng and Yacoob, Yaser and Wang, Lijuan},
  journal={arXiv preprint arXiv:2306.14565},
  year={2023}
}

@article{Favero2024MultiModalHC,
  title={Multi-Modal Hallucination Control by Visual Information Grounding},
  author={Alessandro Favero and Luca Zancato and Matthew Trager and Siddharth Choudhary and Pramuditha Perera and Alessandro Achille and Ashwin Swaminathan and Stefan 0 Soatto},
  journal={2024 IEEE/CVF Conference on Computer Vision and Pattern Recognition (CVPR)},
  year={2024},
  pages={14303-14312},
  url={https://api.semanticscholar.org/CorpusID:268553740}
}

@misc{liu2023llava,
      title={Visual Instruction Tuning}, 
      author={Liu, Haotian and Li, Chunyuan and Wu, Qingyang and Lee, Yong Jae},
      publisher={NeurIPS},
      year={2023},
}

@misc{garg2024imageinwords,
      title={ImageInWords: Unlocking Hyper-Detailed Image Descriptions}, 
      author={Roopal Garg and Andrea Burns and Burcu Karagol Ayan and Yonatan Bitton and Ceslee Montgomery and Yasumasa Onoe and Andrew Bunner and Ranjay Krishna and Jason Baldridge and Radu Soricut},
      year={2024},
      eprint={2405.02793},
      archivePrefix={arXiv},
      primaryClass={cs.CV}
}

@article{lee2024toward,
  title={Toward Robust Hyper-Detailed Image Captioning: A Multiagent Approach and Dual Evaluation Metrics for Factuality and Coverage},
  author={Lee, Saehyung and Yoon, Seunghyun and Bui, Trung and Shi, Jing and Yoon, Sungroh},
  journal={arXiv preprint arXiv:2412.15484},
  year={2024}
}

@inproceedings{
asai2024selfrag,
author={Asai, Akari and Wu, Zeqiu and Wang, Yizhong and Sil, Avirup and Hajishirzi, Hannaneh},
title={Self-{RAG}: Learning to Retrieve, Generate, and Critique through Self-Reflection},
booktitle={The Twelfth International Conference on Learning Representations},
year={2024},
url={https://openreview.net/forum?id=hSyW5go0v8}
}

@article{arora2024just,
  title={Just read twice: closing the recall gap for recurrent language models},
  author={Arora, Simran and Timalsina, Aman and Singhal, Aaryan and Eyuboglu, Sabri and Zhao, Xinyi and Rao, Ashish and Rudra, Atri and Ré, Christopher},
  year={2024}
}

@inproceedings{petroni-etal-2021-kilt,
    title = "{KILT}: a Benchmark for Knowledge Intensive Language Tasks",
    author = {Petroni, Fabio  and Piktus, Aleksandra  and
      Fan, Angela  and Lewis, Patrick  and
      Yazdani, Majid  and De Cao, Nicola  and
      Thorne, James  and Jernite, Yacine  and
      Karpukhin, Vladimir  and Maillard, Jean  and
      Plachouras, Vassilis  and Rockt{\"a}schel, Tim  and
      Riedel, Sebastian},
    booktitle = "Proceedings of the 2021 Conference of the North American Chapter of the Association for Computational Linguistics: Human Language Technologies",
    month = jun,
    year = "2021",
}

@article{kwiatkowski-etal-2019-natural,
    title = "Natural Questions: A Benchmark for Question Answering Research",
    author = "Kwiatkowski, Tom  and
      Palomaki, Jennimaria  and
      Redfield, Olivia  and
      Collins, Michael  and
      Parikh, Ankur  and
      Alberti, Chris  and
      Epstein, Danielle  and
      Polosukhin, Illia  and
      Devlin, Jacob  and
      Lee, Kenton  and
      Toutanova, Kristina  and
      Jones, Llion  and
      Kelcey, Matthew  and
      Chang, Ming-Wei  and
      Dai, Andrew M.  and
      Uszkoreit, Jakob  and
      Le, Quoc  and
      Petrov, Slav",
    journal = "Transactions of the Association for Computational Linguistics",
    year = "2019",
}

@article{2017arXivtriviaqa,
       author = {{Joshi}, Mandar and {Choi}, Eunsol and {Weld},
                 Daniel and {Zettlemoyer}, Luke},
        title = "{triviaqa: A Large Scale Distantly Supervised Challenge Dataset for Reading Comprehension}",
      journal = {arXiv e-prints},
         year = 2017,
          eid = {arXiv:1705.03551},
        pages = {arXiv:1705.03551},
archivePrefix = {arXiv},
       eprint = {1705.03551},
}

@article{levy2017zero,
  title={Zero-shot relation extraction via reading comprehension},
  author={Levy, Omer and Seo, Minjoon and Choi, Eunsol and Zettlemoyer, Luke},
  journal={arXiv preprint arXiv:1706.04115},
  year={2017}
}

@inproceedings{elsahar-etal-2018-rex,
    title = "{T}-{RE}x: A Large Scale Alignment of Natural Language with Knowledge Base Triples",
    author = "Elsahar, Hady  and
      Vougiouklis, Pavlos  and
      Remaci, Arslen  and
      Gravier, Christophe  and
      Hare, Jonathon  and
      Laforest, Frederique  and
      Simperl, Elena",
    booktitle = "Proceedings of the Eleventh International Conference on Language Resources and Evaluation ({LREC} 2018)",
    year = "2018",
}

@article{yang2018hotpotqa,
  title={HotpotQA: A dataset for diverse, explainable multi-hop question answering},
  author={Yang, Zhilin and Qi, Peng and Zhang, Saizheng and Bengio, Yoshua and Cohen, William W and Salakhutdinov, Ruslan and Manning, Christopher D},
  journal={EMNLP},
  year={2018}
}

@article{dua2019drop,
  title={DROP: A reading comprehension benchmark requiring discrete reasoning over paragraphs},
  author={Dua, Dheeru and Wang, Yizhong and Dasigi, Pradeep and Stanovsky, Gabriel and Singh, Sameer and Gardner, Matt},
  journal={arXiv preprint arXiv:1903.00161},
  year={2019}
}

@article{rajpurkar2016squad,
  title={Squad: 100,000+ questions for machine comprehension of text},
  author={Rajpurkar, Pranav and Zhang, Jian and Lopyrev, Konstantin and Liang, Percy},
  journal={EMNLP},
  year={2016}
}

@misc{arora2023language,
      title={Language Models Enable Simple Systems for Generating Structured Views of Heterogeneous Data Lakes},
      author={Simran Arora and Brandon Yang and Sabri Eyuboglu and Avanika Narayan and Andrew Hojel and Immanuel Trummer and Christopher Ré},
      year={2023},
      eprint={2304.09433},
      archivePrefix={arXiv},
      primaryClass={cs.CL}
}

@inproceedings{lockard-etal-2019-openceres,
    title = "{O}pen{C}eres: {W}hen Open Information Extraction Meets the Semi-Structured Web",
    author = "Lockard, Colin  and
      Shiralkar, Prashant  and
      Dong, Xin Luna",
    booktitle = "Proceedings of the 2019 Conference of the North {A}merican Chapter of the Association for Computational Linguistics: Human Language Technologies, Volume 1 (Long and Short Papers)",
    year = "2019",
}

@article{zhou2023context,
  title={Context-faithful Prompting for Large Language Models},
  author={Zhou, Wenxuan and Zhang, Sheng and Poon, Hoifung and Chen, Muhao},
  journal={arXiv preprint arXiv:2303.11315},
  year={2023}
}

@article{touvron2023llama,
  title={Llama 2: Open foundation and fine-tuned chat models},
  author={Touvron, Hugo and Martin, Louis and Stone, Kevin and Albert, Peter and Almahairi, Amjad and Babaei, Yasmine and Bashlykov, Nikolay and Batra, Soumya and Bhargava, Prajjwal and Bhosale, Shruti and others},
  journal={arXiv preprint arXiv:2307.09288},
  year={2023}
}

@article{grattafiori2024llama,
  title={The llama 3 herd of models},
  author={Grattafiori, Aaron and Dubey, Abhimanyu and Jauhri, Abhinav and Pandey, Abhinav and Kadian, Abhishek and Al-Dahle, Ahmad and Letman, Aiesha and Mathur, Akhil and Schelten, Alan and Vaughan, Alex and others},
  journal={arXiv preprint arXiv:2407.21783},
  year={2024}
}

@article{yang2024qwen2,
  title={Qwen2. 5 technical report},
  author={Yang, An and Yang, Baosong and Zhang, Beichen and Hui, Binyuan and Zheng, Bo and Yu, Bowen and Li, Chengyuan and Liu, Dayiheng and Huang, Fei and Wei, Haoran and others},
  journal={arXiv preprint arXiv:2412.15115},
  year={2024}
}

@misc{izacard2021contriever,
      title={Unsupervised Dense Information Retrieval with Contrastive Learning}, 
      author={Gautier Izacard and Mathilde Caron and Lucas Hosseini and Sebastian Riedel and Piotr Bojanowski and Armand Joulin and Edouard Grave},
      year={2021},
      url = {https://arxiv.org/abs/2112.09118},
      doi = {10.48550/ARXIV.2112.09118},
}

@inproceedings{corg,
    title = "{CORG}: Generating Answers from Complex, Interrelated Contexts",
    author = "Lee, Hyunji  and
      Dernoncourt, Franck  and
      Bui, Trung  and
      Yoon, Seunghyun",
    booktitle = "Proceedings of the 2025 Conference of the Nations of the Americas Chapter of the Association for Computational Linguistics: Human Language Technologies (Volume 1: Long Papers)",
    month = apr,
    year = "2025",
}

@article{chen2024far,
  title={How far are we to gpt-4v? closing the gap to commercial multimodal models with open-source suites},
  author={Chen, Zhe and Wang, Weiyun and Tian, Hao and Ye, Shenglong and Gao, Zhangwei and Cui, Erfei and Tong, Wenwen and Hu, Kongzhi and Luo, Jiapeng and Ma, Zheng and others},
  journal={Science China Information Sciences},
  volume={67},
  number={12},
  pages={220101},
  year={2024},
  publisher={Springer}
}

@inproceedings{lu2022learn,
    title={Learn to Explain: Multimodal Reasoning via Thought Chains for Science Question Answering},
    author={Lu, Pan and Mishra, Swaroop and Xia, Tony and Qiu, Liang and Chang, Kai-Wei and Zhu, Song-Chun and Tafjord, Oyvind and Clark, Peter and Ashwin Kalyan},
    booktitle={The 36th Conference on Neural Information Processing Systems (NeurIPS)},
    year={2022}
}

@article{MMBench,
    author  = {Liu, Yuan and Duan, Haodong and Zhang, Yuanhan and Li, Bo and Zhang, Songyang and Zhao, Wangbo and Yuan, Yike and Wang, Jiaqi and He,  Conghui and Liu, Ziwei and Chen, Kai and Lin, Dahua},
    journal = {arXiv:2307.06281},
    title   = {MMBench: Is Your Multi-modal Model an All-around Player?},
    year    = {2023},
}

@inproceedings{sun2023mme,
  title={Masked Motion Encoding for Self-Supervised Video Representation Learning},
  author={Sun, Xinyu and Chen, Peihao and Chen, Liangwei and Li, Changhao and Li, Thomas H and Tan, Mingkui and Gan, Chuang},
  booktitle={The IEEE/CVF Conference on Computer Vision and Pattern Recognition (CVPR)},
  year={2023}
}

@inproceedings{hudson2019gqa,
  title={Gqa: A new dataset for real-world visual reasoning and compositional question answering},
  author={Hudson, Drew A and Manning, Christopher D},
  booktitle={Proceedings of the IEEE/CVF conference on computer vision and pattern recognition},
  pages={6700--6709},
  year={2019}
}

@inproceedings{Bisk2020,
    author = {Yonatan Bisk and Rowan Zellers and
            Ronan Le Bras and Jianfeng Gao
            and Yejin Choi},
    title = {PIQA: Reasoning about Physical Commonsense in
           Natural Language},
    booktitle = {Thirty-Fourth AAAI Conference on
               Artificial Intelligence},
    year = {2020},
}

@inproceedings{radford2021learning,
  title={Learning transferable visual models from natural language supervision},
  author={Radford, Alec and Kim, Jong Wook and Hallacy, Chris and Ramesh, Aditya and Goh, Gabriel and Agarwal, Sandhini and Sastry, Girish and Askell, Amanda and Mishkin, Pamela and Clark, Jack and others},
  booktitle={International conference on machine learning},
  pages={8748--8763},
  year={2021},
  organization={PmLR}
}

@inproceedings{sap2019social,
  title={Social IQa: Commonsense Reasoning about Social Interactions},
  author={Sap, Maarten and Rashkin, Hannah and Chen, Derek and Le Bras, Ronan and Choi, Yejin},
  booktitle={Proceedings of the 2019 Conference on Empirical Methods in Natural Language Processing and the 9th International Joint Conference on Natural Language Processing (EMNLP-IJCNLP)},
  pages={4463--4473},
  year={2019}
}

@article{sakaguchi2019winogrande,
    title={WinoGrande: An Adversarial Winograd Schema Challenge at Scale},
    author={Sakaguchi, Keisuke and Bras, Ronan Le and Bhagavatula, Chandra and Choi, Yejin},
    journal={arXiv preprint arXiv:1907.10641},
    year={2019}
}

@inproceedings{zellers2019hellaswag,
    title={HellaSwag: Can a Machine Really Finish Your Sentence?},
    author={Zellers, Rowan and Holtzman, Ari and Bisk, Yonatan and Farhadi, Ali and Choi, Yejin},
    booktitle ={Proceedings of the 57th Annual Meeting of the Association for Computational Linguistics},
    year={2019}
}

@misc{lambada, 
    author={Paperno, Denis and Kruszewski, Germán and Lazaridou, Angeliki and Pham, Quan Ngoc and Bernardi, Raffaella and Pezzelle, Sandro and Baroni, Marco and Boleda, Gemma and Fernández, Raquel}, 
    title={The LAMBADA dataset}, 
    DOI={10.5281/zenodo.2630551}, 
    publisher={Zenodo}, 
    year={2016}, 
    month={Aug} 
}

@article{Clark2018ThinkYH,
  title={Think you have Solved Question Answering? Try ARC, the AI2 Reasoning Challenge},
  author={Peter Clark and Isaac Cowhey and Oren Etzioni and Tushar Khot and Ashish Sabharwal and Carissa Schoenick and Oyvind Tafjord},
  journal={ArXiv},
  year={2018},
  volume={abs/1803.05457}
}

@article{2023llavarlhf,
    author      = {Zhiqing Sun and Sheng Shen and Shengcao Cao and Haotian Liu and Chunyuan Li and Yikang Shen and Chuang Gan and Liang-Yan Gui and Yu-Xiong Wang and Yiming Yang and Kurt Keutzer and Trevor Darrell},
    title       = {Aligning Large Multimodal Models with Factually Augmented RLHF},
    publisher   = {arXiv:2309.14525},
    year        = {2023}
}

@misc{eval-harness,
  author       = {Gao, Leo and Tow, Jonathan and Abbasi, Baber and Biderman, Stella and Black, Sid and DiPofi, Anthony and Foster, Charles and Golding, Laurence and Hsu, Jeffrey and Le Noac'h, Alain and Li, Haonan and McDonell, Kyle and Muennighoff, Niklas and Ociepa, Chris and Phang, Jason and Reynolds, Laria and Schoelkopf, Hailey and Skowron, Aviya and Sutawika, Lintang and Tang, Eric and Thite, Anish and Wang, Ben and Wang, Kevin and Zou, Andy},
  title        = {The Language Model Evaluation Harness},
  month        = 07,
  year         = 2024,
  publisher    = {Zenodo},
  version      = {v0.4.3},
  doi          = {10.5281/zenodo.12608602},
  url          = {https://zenodo.org/records/12608602}
}

@article{goyal2024context,
  title={Context-parametric inversion: Why instruction finetuning can worsen context reliance},
  author={Goyal, Sachin and Baek, Christina and Kolter, J Zico and Raghunathan, Aditi},
  journal={arXiv preprint arXiv:2410.10796},
  year={2024}
}

@article{luo2023sail,
  title={Sail: Search-augmented instruction learning},
  author={Luo, Hongyin and Chuang, Yung-Sung and Gong, Yuan and Zhang, Tianhua and Kim, Yoon and Wu, Xixin and Fox, Danny and Meng, Helen and Glass, James},
  journal={arXiv preprint arXiv:2305.15225},
  year={2023}
}

@article{wang2024loss,
  title={On the loss of context-awareness in general instruction fine-tuning},
  author={Wang, Yihan and Bai, Andrew and Peng, Nanyun and Hsieh, Cho-Jui},
  journal={arXiv preprint arXiv:2411.02688},
  year={2024}
}

@article{an2024make,
  title={Make your llm fully utilize the context},
  author={An, Shengnan and Ma, Zexiong and Lin, Zeqi and Zheng, Nanning and Lou, Jian-Guang and Chen, Weizhu},
  journal={Advances in Neural Information Processing Systems},
  volume={37},
  pages={62160--62188},
  year={2024}
}

@article{wu2023language,
  title={From language modeling to instruction following: Understanding the behavior shift in llms after instruction tuning},
  author={Wu, Xuansheng and Yao, Wenlin and Chen, Jianshu and Pan, Xiaoman and Wang, Xiaoyang and Liu, Ninghao and Yu, Dong},
  journal={arXiv preprint arXiv:2310.00492},
  year={2023}
}

@article{begin2025pause,
  title={Pause-Tuning for Long-Context Comprehension: A Lightweight Approach to LLM Attention Recalibration},
  author={Begin, James and Agrawal, Namit and Singh, Eshan and Fu, Yicheng and O'Brien, Sean and Sharma, Vasu and Zhu, Kevin},
  journal={arXiv preprint arXiv:2502.20405},
  year={2025}
}

@article{he2023never,
  title={Never lost in the middle: Mastering long-context question answering with position-agnostic decompositional training},
  author={He, Junqing and Pan, Kunhao and Dong, Xiaoqun and Song, Zhuoyang and Liu, Yibo and Sun, Qianguo and Liang, Yuxin and Wang, Hao and Zhang, Enming and Zhang, Jiaxing},
  journal={arXiv preprint arXiv:2311.09198},
  year={2023}
}

@article{liu2023lost,
  title={Lost in the middle: How language models use long contexts},
  author={Liu, Nelson F and Lin, Kevin and Hewitt, John and Paranjape, Ashwin and Bevilacqua, Michele and Petroni, Fabio and Liang, Percy},
  journal={arXiv preprint arXiv:2307.03172},
  year={2023}
}

@article{gao2023roles,
  title={Roles of scaling and instruction tuning in language perception: Model vs. human attention},
  author={Gao, Changjiang and Huang, Shujian and Li, Jixing and Chen, Jiajun},
  journal={arXiv preprint arXiv:2310.19084},
  year={2023}
}

@online{DatabricksBlog2023DollyV2,
    author    = {Mike Conover and Matt Hayes and Ankit Mathur and Jianwei Xie and Jun Wan and Sam Shah and Ali Ghodsi and Patrick Wendell and Matei Zaharia and Reynold Xin},
    title     = {Free Dolly: Introducing the World's First Truly Open Instruction-Tuned LLM},
    year      = {2023},
    url       = {https://www.databricks.com/blog/2023/04/12/dolly-first-open-commercially-viable-instruction-tuned-llm},
    urldate   = {2023-06-30}
}

@article{wei2021finetuned,
  title={Finetuned language models are zero-shot learners},
  author={Wei, Jason and Bosma, Maarten and Zhao, Vincent Y and Guu, Kelvin and Yu, Adams Wei and Lester, Brian and Du, Nan and Dai, Andrew M and Le, Quoc V},
  journal={arXiv preprint arXiv:2109.01652},
  year={2021}
}

@article{ouyang2022training,
  title={Training language models to follow instructions with human feedback},
  author={Ouyang, Long and Wu, Jeffrey and Jiang, Xu and Almeida, Diogo and Wainwright, Carroll and Mishkin, Pamela and Zhang, Chong and Agarwal, Sandhini and Slama, Katarina and Ray, Alex and others},
  journal={Advances in neural information processing systems},
  volume={35},
  pages={27730--27744},
  year={2022}
}

@article{sanh2021multitask,
  title={Multitask prompted training enables zero-shot task generalization},
  author={Sanh, Victor and Webson, Albert and Raffel, Colin and Bach, Stephen H and Sutawika, Lintang and Alyafeai, Zaid and Chaffin, Antoine and Stiegler, Arnaud and Scao, Teven Le and Raja, Arun and others},
  journal={arXiv preprint arXiv:2110.08207},
  year={2021}
}

@misc{alpaca,
  author = {Rohan Taori and Ishaan Gulrajani and Tianyi Zhang and Yann Dubois and Xuechen Li and Carlos Guestrin and Percy Liang and Tatsunori B. Hashimoto },
  title = {Stanford Alpaca: An Instruction-following LLaMA model},
  year = {2023},
  publisher = {GitHub},
  journal = {GitHub repository},
  howpublished = {\url{https://github.com/tatsu-lab/stanford_alpaca}},
}

\appendix

\begin{table*}[t!]
\centering
\fontsize{7}{10} \selectfont
    \begin{tabular}{ccc|ccccc|cc|cccc|c}
    \toprule
    Ctx Presence & Ctx Placement & Ctx Loss & NQ & TQA & zsRE & T-rex & HQA &  NQ-C & Corg  & Drop & Squad & SWDE & FDA & \textbf{\textit{Avg}} \\ 
    \midrule
    F & - & - & \text{48.0} & 70.2& 57.5& 62.7 & 48.0 & 60.7 & 18.0 & 39.0 & 64.9 & 92.0 & 74.0 & 57.7 \\ 
    T & Input & T & 42.9& 69.0& 60.2 &57.9 & 43.5 & 33.6 & 11.2 &18.0 & 54.5 & 90.8& 71.2 & 50.3 \\ 
    T & Input & F & 47.1 & 72.0 & 60.3 & 62.9 & 49.5 & 72.1 & 20.6 & 45.1 & 68.8 & 95.7 & 78.9 & 61.2\\
    T & Output & T  &45.8  & 71.7 & 60.0 & \text{64.1} &48.1 & 47.0 & 13.1& 30.3 &63.8 & 94.3 & 76.9 & 55.9\\
    T & Output & F  & \text{46.2} & \text{72.6} & \text{62.6} & \text{63.4} & \text{50.6} & \text{72.3} & \text{19.4} & \text{44.0} & \text{69.5} & \text{95.0} & \text{80.9} & \textbf{61.5}\\ 
    \bottomrule
    \end{tabular}
\caption
     {
     Performance comparison over various context choices; Ctx Presence is whether the context is added to the response. Ctx Placement is whether the relevant context is added to the input (instruction) or output (before each corresponding response sentence). Ctx Loss is rather we calculate loss over the context or not.
     We train all models using Llama3.1 8B as base model and evaluate over 11 information-seeking datasets. 
     } 
\label{table: full}
\end{table*}

\section{Experimental Setup} \label{app_sec: exp_setup}
\subsection{Comparison Models: \wcontext vs. \wocontext}
\label{app_subsec: model_comparison}
Table~\ref{table: full} shows performance of various models trained on various design choices of provided context. There are three axes: whether the context exists in training dataset (Ctx Presence), whether the context is added to the input (instruction) or output (next to the corresponding response sentences) (Ctx Placement), or whether to calculate loss over the context or not (Ctx Loss).
For case where context exists, we used the case where context is added to the output side with no loss over the context following previous works~\citep{lee2024semiparametric, asai2024selfrag}. Also this showed highest performance.
We also observed some interesting analysis by the choices when context is present (\wcontext): 

\paragraph{Ctx Placement}

When comparing performance on model when training with context added to the input (instruction) or output (assistant response), we observe that overall performance is comparable across both settings, with a slight average improvement when context is added to the output, aligning with findings from prior work~\citep{lee2024semiparametric, asai2024selfrag}, which suggest that placing context closer to the generation target improves grounding performance.
Main difference for these two comes when there are multiple evidences for single example, thereby rather to concatenate all and add to user prompt or to separate each into relevant response sentence and place them in front.

\begin{figure}[t]
\centering
\includegraphics[width=\linewidth]{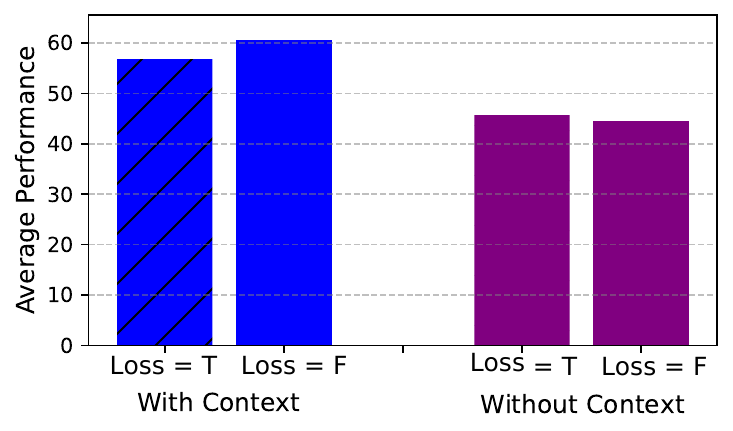}
\caption{
    Performance of models trained with context, either with loss over the context (Loss = T) or without loss over the context (Loss = F). \textit{With Context} indicates inference with additional context provided, while \textit{Without Context} indicates inference without additional context.
    }
\label{fig_app: with_without_context_styled}
\end{figure}

\paragraph{Ctx Loss}

We experiment over two training settings: one where the loss is computed over both the context and the response (loss=T), and one where it is computed only over the response (loss=F).
As shown in Figure~\ref{fig_app: with_without_context_styled}, models trained without loss on the context (loss=F) perform better when context is available at inference. However, their performance drops to the same level or worse than model trained with loss on the context (loss=T) when no context is provided, thereby the model must rely on its parametric knowledge.

We hypothesize that this occurs because, under loss=T, the model must encode and store the given context in its parameters to minimize training loss, which later helps when no external context is available. 
In contrast, models trained with loss=F learn how to use provided context at inference rather than memorizing it, as also observed in Section~\ref{behavior: attn}.

\subsection{Text Domain} \label{app: exp_text}

\paragraph{Datasets}

For training, we use the 29k dataset from Self-RAG~\citep{asai2024selfrag}, which is constructed by augmenting instruction-tuning datasets with sentence-level relevant context, incorporated when available. Following the filtering procedure from prior work~\citep{lee2024semiparametric}, we retain only instances with relevant context. Additionally, we exclude examples where the generation of counterfactual contexts fails, to facilitate more focused analysis and experimentation on counterfactual behavior (see Appendix~\ref{app_behavior: counterfactual_dataset}).

For evaluation, we experiment over 11 information-seeking datasets, including NQ~\citep{kwiatkowski-etal-2019-natural}, TriviaQA~(TQA)~\citep{2017arXivtriviaqa}, zsRE~\citep{levy2017zero}, T-rex~\citep{elsahar-etal-2018-rex}, and HotpotQA~(HQA)~\citep{yang2018hotpotqa}, using the versions provided in KILT~\citep{petroni-etal-2021-kilt}. 
In all experiments, we keep the external context frozen, as the focus of this work is on evaluating the language model itself. We use either the gold contexts annotated in KILT or the top-20 passages retrieved by contriever-msmarco~\citep{izacard2021contriever}, a strong dense retrieval model.
We also include DROP~\citep{dua2019drop}, SQuAD~\citep{rajpurkar2016squad}, SWDE~\citep{lockard-etal-2019-openceres}, and FDA~\citep{arora2023language}, for which we use the version curated by Based~\citep{arora2024just}. Additionally, we evaluate on two benchmarks specifically designed to highlight grounding failures in language models: NQ Conflict (NQ-C)~\citep{zhou2023context} and the dataset from CORG~\citep{corg}\footnote{The datasets used in our experiments are released under the following licenses: Natural Questions (NQ), SWDE, and NQ Conflict (NQ-C) under the Creative Commons Attribution 4.0 (CC BY 4.0) license; TriviaQA (TQA), T-REx, HotpotQA (HQA), DROP, and SQuAD under the Creative Commons Attribution-ShareAlike 4.0 (CC BY-SA 4.0) license; zsRE and CORG under the MIT License; and FDA under the Apache License 2.0.
}. 
For both training and evaluation dataset, we used English dataset.

For the dataset from CORG, we report the D-F1 metric introduced in their work, which measures whether the generated response contains a disambiguated correct answer. For the other datasets, we evaluate using answer accuracy, which measures whether the correct answer appears in the generated response.

We further evaluate with 7 downstream tasks (PIQA~\citep{Bisk2020}, Social IQa~\citep{sap2019social}, Winogrande~\citep{sakaguchi2019winogrande}, HellaSwag~\citep{zellers2019hellaswag}, LAMBADA-OpenAI~\citep{lambada}, ARC-Challenge, and ARC-Easy~\citep{Clark2018ThinkYH}) to evaluate overall language model performance through lm-evaluation-harness~\citep{eval-harness}.
We report normalized answer scores for all tasks except ARC-Easy, for which we use answer accuracy.

\subsection{Visual Language Domain} \label{app: exp_vlm}
\paragraph{Baseline}

We compare vision-language models that use a language model trained on context-free data (\wocontext) versus context-augmented data (\wcontext) as the LLM backbone for vision language alignment, resulting in \wovlm and \wvlm, respectively.
The backbone LLMs are the same as those evaluated in Section~\ref{app: exp_text}.
For all models, we adopt a widely used two-stage vision-language training pipeline~\citep{chen2024far, liu2023llava}: (1) a pretraining stage for feature alignment, and (2) an end-to-end fine-tuning stage. Note that the vision-language alignment procedure is identical across all experiments, and only the backbone language model varies. 

\paragraph{Datasets \& Evaluation Metrics}

When training the vision-language alignment, we use the training dataset from LLaVA~\citep{liu2023llava} for both the pretraining and finetuning stages; filtered CC-595K subset for pretraining and LLaVA-Instruct-158K for finetuning.
To evaluate how well each model grounds its responses to the provided image, we mainly conduct experiments on four benchmarks commonly used to measure hallucination in vision-language models: AMBER~\citep{wang2023llm}, POPE~\citep{li2023evaluating}, ImageInWords~\citep{garg2024imageinwords}, and LLaVA-Wild~\citep{liu2023llava}.
For POPE, we report the average F1 score across all splits (popular, adversarial, and random).
For AMBER, we evaluate performance on both generative and discriminative tasks.
For ImageInWords, we adopt the evaluation metric from CapMAS~\citep{lee2024toward}, which uses a GPT-based method to assess factuality in fine-grained manner.  
For LLaVA-Wild, we use the general GPT4-Eval metric. We adjust the evaluation rubric to explicitly penalize hallucinations, focusing the evaluation more precisely on hallucination detection.
To assess overall model capabilities beyond hallucination, we additionally evaluate on four widely used downstream benchmarks: MMBench~\citep{MMBench}, ScienceQA~\citep{lu2022learn}, MME~\citep{sun2023mme}, and GQA~\citep{hudson2019gqa}.

\paragraph{Training Details}

All models are trained using the same vision-language alignment procedure, with the only difference being the choice of language model backbone. We follow the training setup introduced in LLaVA~\citep{liu2023llava}, and conduct all experiments on 8 NVIDIA A100 GPUs.
During the pretraining stage, the language model is kept frozen and only the projection layer, which maps image features to language model's word embedding space, is trained. The pretraining stage is run for one epoch with a learning rate of 2e-3 and a batch size of 128. In the subsequent fine-tuning stage, both the projection layer and the language model are updated, while the vision encoder remains frozen throughout. Fine-tuning is performed for three epochs using a learning rate of 2e-5 and a batch size of 32. For the vision encoder, we use the pre-trained CLIP visual encoder ViT-L/14~\citep{radford2021learning} following previous works~\citep{liu2023llava, 2023llavarlhf}.

\section{RQ1: How does training an LLM on instruction tuning instruction tuning with instances containing context differ from tuning without context?}

\subsection{Effect of number of contexts on performance}

\begin{figure}[t]
\centering
\includegraphics[width=\linewidth]{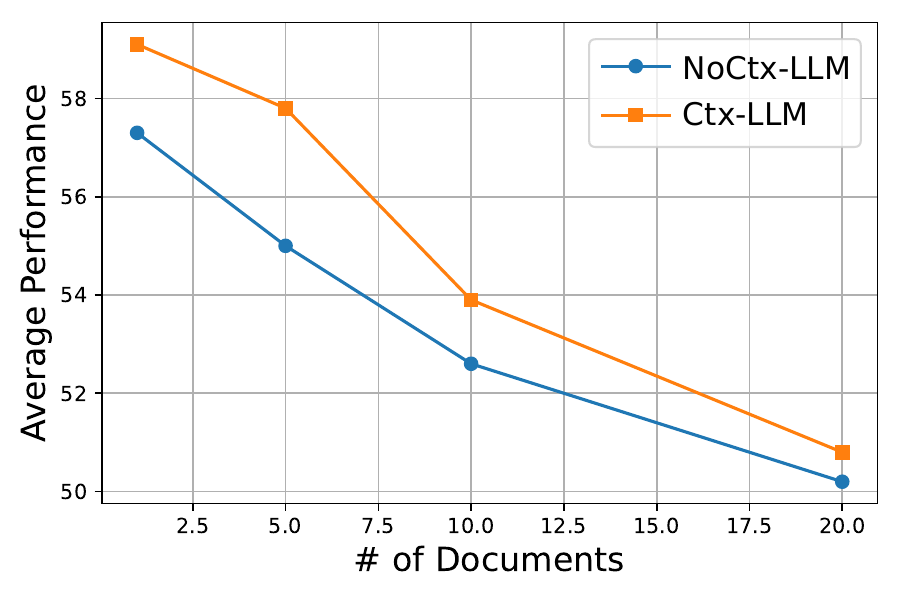}
\caption{
Avg. accuracy across nine datasets (y-axis), NQ, TriviaQA, zsRE, T-REx, and HotpotQA, as a function of the number of contexts for each instance (x-axis), for models trained with context (\wcontext) and without context (\wocontext).
}
\label{fig_app: retrieval_avg_perf}
\end{figure}

Figure~\ref{fig_app: retrieval_avg_perf} shows the average performance across five datasets (NQ, TriviaQA, zsRE, T-REx, and HotpotQA) (y-axis) as the number of retrieved context increases (x-axis) for \wcontext and \wocontext, using LLaMA 3.1 8B as the base model.
We observe that \wcontext consistently outperforms \wocontext, but the performance gap narrows as more contexts are added.
We hypothesize that this is because \wcontext encourages strong grounding to the provided context; therefore, when many potentially distracting contexts are present, the model may become susceptible to being misled by irrelevant information.
\begin{figure*}[t!]
    \centering
    \begin{minipage}[t]{0.95\linewidth}
        \centering
        \includegraphics[width=\linewidth]{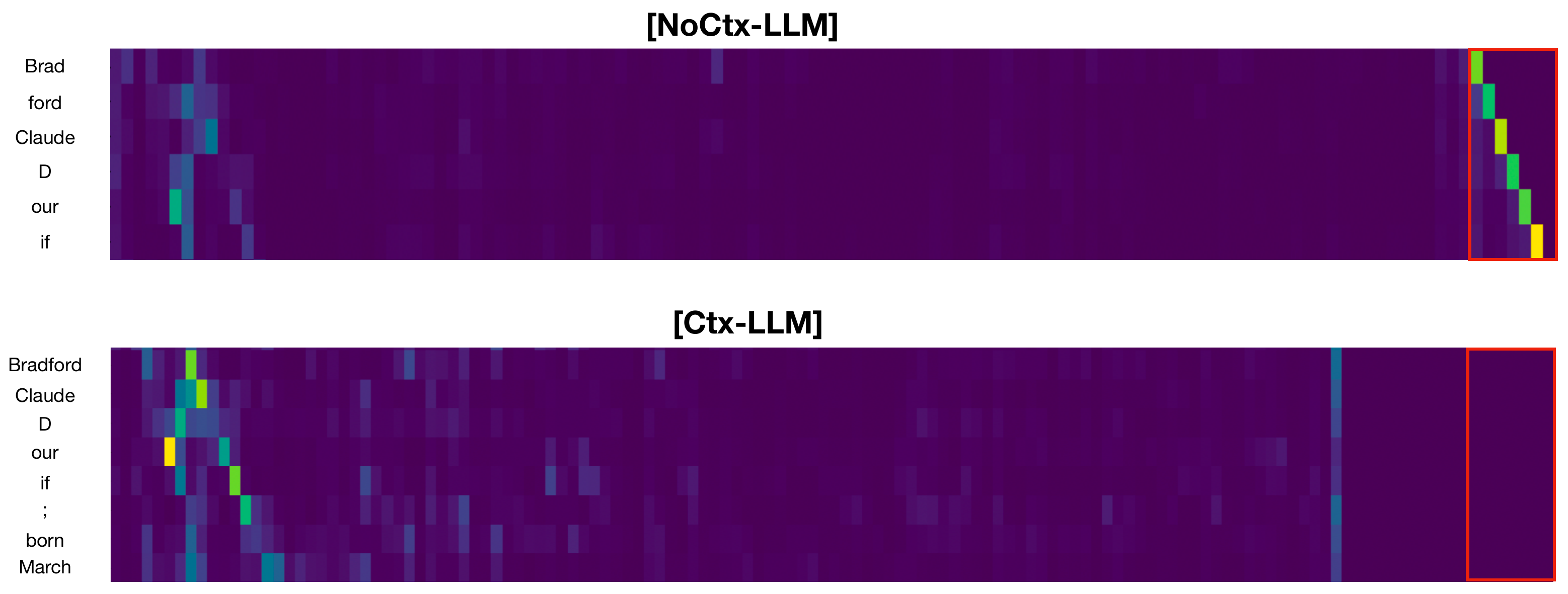}
        \caption{
        Attention maps during response generation for models trained with \wocontext (top) and with \wcontext (bottom)
        }
        \label{fig_app: attn_map}
    \end{minipage}
\end{figure*}

\subsection{Attention Patterns Differ Between \wocontext and \wcontext} \label{app_sec: behavior}
Figure~\ref{fig_app: attn_map} presents attention maps during response generation for \wocontext (top) and \wcontext (bottom).
\wcontext display stronger attention to the input context, whereas \wocontext attend more heavily to previously generated tokens (highlighted in the red box).
This suggests that training with context-augmented data (\wcontext) encourages models to remain more grounded in the input, rather than relying on self-generated content.

\subsection{Counterfactual Dataset Construction} \label{app_behavior: counterfactual_dataset}
In this section, we describe our procedure for constructing the counterfactual dataset, which are datasets that contains knowledge that counterfacts with model's prior knowledge.

\paragraph{Dataset Construction \& Validation}
We categorize the dataset into two groups based on the format of the original answers: True/False (T/F) and Free-Form.
For T/F examples, where the answer is either true or false, we generate counterfactuals by simply inverting the original boolean value.
For Free-Form examples, which comprise the remainder of the dataset, we prompt the model to generate a counterfactual response using the template shown in Figure~\ref{fig:fake_answer}.
We discard the samples for which GPT-4o\footnote{\url{https://openai.com/index/hello-gpt-4o/}} refuses generation or fails to match the required format.

Using the generated counterfactual answer, we then ask GPT-4o (in the same session) to fabricate a supporting “background reference” using the template in Figure~\ref{fig:fake_context}.

To ensure that each fake pair (generated counterfactual answer and external-knowledge) remains coherent, we re-initialize a fresh GPT-4o chat and validate with the template in Figure~\ref{fig:validation}.
Only samples for which GPT-4o selects \texttt{Answer: B} are retained.

\paragraph{Evaluation Dataset Construction}
We randomly sample 1k examples from the dataset and use GPT-5 to generate cloze-style questions, which are then used to evaluate whether the model knows the corresponding knowledge. Template we used to construct the evaluation dataset is in Figure~\ref{fig:clozed_eval}.

\paragraph{Dataset Statistics and Cost}
The overall validation pass rates are: \textbf{T/F}: 95.5\%; \textbf{Free-Form}: 97.9\%.
In total, we obtain 29k validated counterfactual examples (Free-Form: 24k; T/F: 5k), at a generation cost of USD \$77.83.

\begin{figure*}[h]
\begin{tcolorbox}[width=0.99\textwidth, halign title=center, title = {Fake Answer Generation Prompt Template}]
You are tasked to create a binary-choice question by creating an alternative wrong answer to the provided question.\newline
\newline
Query: \{query\}\newline
\newline
Ground Truth Answer: \{answer\}\newline
\newline
Create a plausible wrong answer for the provided question. Your response should be in the format of the following: \newline
\newline
Wrong Answer: \(<\)Plausible Wrong Answer\(>\)\newline
\end{tcolorbox}
\caption{Template for Generating Fake Answers.}
\label{fig:fake_answer}
\end{figure*}

\begin{figure*}[h]
\begin{tcolorbox}[width=0.99\textwidth, halign title=center, title = {Fake External-Knowledge Generation Prompt Template}]
Now, create a background reference from Wikipedia that supports your generated wrong answer. Keep the length of the reference around 100 words. Remember, your generated fictional reference should be convincing as possible so that people will be tempted to choose your generated wrong answer, **instead of the original ground truth answer!** The generated reference passage should seem like an excerpt from Wikipedia. This means that the reference passage should NOT start with 'According to ...'. You must NOT mention the original answer in your new reference passage. Answer in the format of the following:\newline
\newline
Reference Passage (Around 100 Words): \(<\)Fictional Passage\(>\)
\end{tcolorbox}
\caption{Template for Generating Fake External-Knowledge.}
\label{fig:fake_context}
\end{figure*}

\begin{figure*}[h]
\begin{tcolorbox}[width=0.99\textwidth, halign title=center, title = {Validation Prompt Template}]
For the following query, pick the answer which is directly supported by the provided passage from Wikipedia. The correctness of the answer does not matter. Focus only on which answer the reference passage directly supports.\newline
\newline
Query: \{query\}\newline
\newline
Reference Passage: \{fake\_reference\}\newline
\newline
Choice of Answers:\newline
A: \{original\_answer\}\newline
B: \{fake\_answer\}\newline
\newline
Provide your answer in the following format:\newline
\newline
Answer: \(<\)A or B\(>\)
\end{tcolorbox}
\caption{Template for Validation.}
\label{fig:validation}
\end{figure*}

\begin{figure*}[h]
\begin{tcolorbox}[width=0.99\textwidth, halign title=center, title = {Cloze-Style Question Generation}]
You are given a passage and a target answer.
Your task is to create a cloze-style questions where the blank should be filled by the given answer and placed at the end of the sentence. Avoid simply copying the passage verbatim; rewrite or paraphrase so it feels like a natural question.
\newline 
\newline 
Context: \{context\}\newline 
\newline 
Ground Truth Answer: \{answer\}\newline
\newline
\end{tcolorbox}
\caption{Template for Generating Cloze-style Questions.}
\label{fig:clozed_eval}
\end{figure*}

\paragraph{Results}
As shown in Figure~\ref{fig:heatmap}, NoCtx(CF) shows a large drop when tested on original knowledge (Ori–Ctx) but a strong improvement when tested on counterfactual knowledge (CF–Ctx), indicating greater forgetting of prior knowledge and strong memorization of the updated (counterfactual) training signal. 
In contrast, models trained with context retain more of their original knowledge but memorizes counterfactual knowledge less strongly: Ctx(CF) outperforms NoCtx(CF) on Ori–Ctx, but scoring lower on CF–Ctx. 
Ctx models also consistently achieve higher performance than NoCtx models whenever context is available (*+Ctx), regardless of whether the context supports original or counterfactual answers.
These findings suggest that context-augmented training shifts the model’s reliance from parametric memory toward external context.

\section{RQ2: How does using \wcontext or \wocontext as the backbone for vision-language adaptation influence performance on vision-language tasks?}

\subsection{\wvlm show robust performance across varying response lengths} \label{app_sec: response_len}
\begin{figure*}[t!]
    \centering
    \begin{minipage}[t]{\linewidth}
        \centering
        \includegraphics[width=0.95\linewidth]{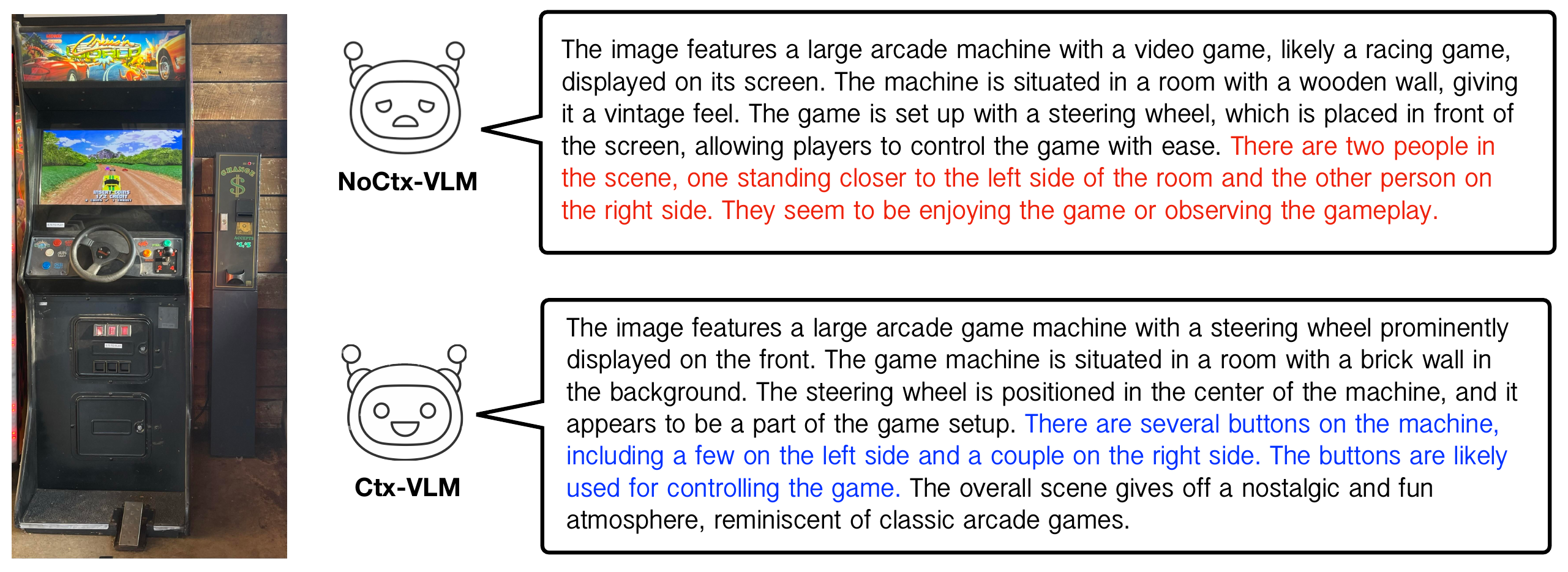}
        \caption{Example from ImageInWords. Fine-grained captions generated for the figure on the left by \wovlm and \wvlm.}
        \label{fig: qualitative}
    \end{minipage}
\end{figure*}

Figure~\ref{fig: qualitative} presents a qualitative comparison of fine-grained captions generated by \wovlm and \wvlm for an instance from the ImageInWords dataset.
The \wovlm tends to hallucinate, incorrectly stating that ``there are two people in the scene'' (red text), likely due to internal priors about humans typically playing arcade games. 
In contrast, \wvlm avoids this hallucination and instead provides a more grounded and descriptive observation (blue text), accurately noting the presence of specific buttons on the arcade machine. We observe similar trends across various other examples.

\paragraph{Effect of Rubric Modification on LLaVA-Wild Performance}
To better assess whether the model hallucinates, we modify the evaluation rubric to impose an additional penalty for hallucinated content: “Responses must remain grounded in the input image. Any hallucinated details should be heavily penalized.” With this revised rubric, we observe that the performance gap between the \wovlm and \wvlm increases significantly: from an average difference of 8.8\% under the original rubric to 18.1\% with the modified one. This suggests that a substantial portion of the performance improvement with \wvlm compared to \wovlm comes from reducing hallucinations; \wvlm generates more factual, image-grounded responses.

\begin{table*}[t!]
\centering
\caption
     {
     Performance on VLM hallucination benchmarks using LLMs trained under different context configurations during instruction tuning.
     } 
\fontsize{7}{10} \selectfont
    \begin{tabular}{cc|c|cccccc|c|ccc}
    \toprule
     \multicolumn{2}{c}{Ctx Configuration}& Pope & \multicolumn{6}{c}{Amber} & Llava-W & \multicolumn{1}{c}{ImageInWords} \\
    \midrule
    Ctx Presence & Ctx Loss & F1 & CHAIR~($\downarrow$) & Cover~($\uparrow$) & Hal~($\uparrow$) & Cog~($\downarrow$) & F1~($\uparrow$) & AMBER~($\uparrow$) &GPT4-Eval & F1 \\
    \midrule
    F & - &87.3 & 9.1& 53.8&39.8  &5.9  & 66.5 & 78.7 & 55.7 & 54.5  \\
    T & T & 87.3 &\textbf{8.6} &53.4 & 37.0& \textbf{4.8} & 70.6 & 81.0 & 62.7 &  55.1 \\
    T & F& \textbf{87.7} &  \textbf{8.6} & \textbf{54.6} &  \textbf{47.2} & 4.9 & \textbf{71.7} & \textbf{81.6}& \textbf{74.2} & \textbf{56.8} \\
    \bottomrule
    \end{tabular}

\label{table_app: vlm_hallucination_sup}
\end{table*}

\paragraph{Effect of context configuration in LLM instruction tuning on vision-language hallucination}
\begin{table}[t!]
\centering
\caption
     {
     Performance difference between using the original Llama-Wild rubric (Original) and a modified version that imposes stronger penalties for hallucinations (Changed).
     } 
\fontsize{7}{10} \selectfont
    \begin{tabular}{c|cc|cc|cc}
    \toprule
     & \multicolumn{2}{c}{Llama2 (7B)} & \multicolumn{2}{c}{Llama3.1 (8B)} & \multicolumn{2}{c}{Qwen2.5 (7B)} \\
    \midrule
    Original & 66.3 & 53.4 & 67.3 & 55.7 & 69.6 & 60.1 \\
    Changed & 72.1 & 70.9 & 77.1 & 74.2 & 80.1 & 78.5 \\
    \bottomrule
    \end{tabular}

\label{table_app: llava-wild}
\end{table}
We evaluate the effect of context configuration during LLM instruction tuning on vision-language adaptation.
Building on the models trained on various context configurations in Section~\ref{app_subsec: model_comparison}, we use these LLMs as backbones for the vision-language alignment and assess their performance on vision-language hallucination benchmarks.

As shown in Table~\ref{table_app: vlm_hallucination_sup}, VLM using LLM trained without computing loss on the context (Ctx Presence = T, Ctx Loss = F) as backbone consistently outperforms both VLM using LLMs trained with loss on the context (Ctx Presence = T, Ctx Loss = T) and those trained without any added context (Ctx Presence = F) as backbone.
This pattern mirrors our findings in the text domain (Section~\ref{app_subsec: model_comparison}) and provides evidence that the context configuration used during instruction tuning influences how models utilize and source knowledge, ultimately affecting downstream performance.

\begin{table*}[t!]
\centering
\caption
     {
     Performance across 11 information-seeking datasets using models trained with \wcontext and \wocontext, applied via LoRA on the LLaMA 3.1 8B model.
     } 
\fontsize{7}{10} \selectfont
    \begin{tabular}{c|ccccc|cc|cccc}
    \toprule
     Training & NQ & TQA & zsRE & T-rex & HotpotQA & NQ-C & Corg & Drop & Squad & SWDE & FDA \\ 
    \midrule
    \wocontext & 41.9 & 63.1 & 53.0 & 58.1 & 39.8 & 49.9 & 10.2 & 35.1 & 57.3 & 89.1 & 70.3 \\ 
    \wcontext & 44.2 & 66.8 & 57.6 & 57.9 & 43.1 & 64.1 & 13.7 & 40.6 & 55.9 & 92.7 & 73.2\\ 
    \bottomrule
    \end{tabular}

\label{table_app: Lora}
\end{table*}

\section{RQ3: How can these insights guide when to use each model and combine them effectively for downstream applications?}

\subsection{Using \wcontext over \wocontext is complementary with inference-time grounding techniques} \label{app_sub: inference_time_grounding}

\begin{table*}[t!]
\centering
\caption
    {
    Performance over 11 information-seeking datasets using inference-based methods with \wocontext and \wcontext trained with Llama 2 7B or Llama 3.1 8B as base models. TQA and HQA refer to TriviaQA and HotpotQA, respectively. 
     } 
\fontsize{7}{10} \selectfont
    \begin{tabular}{c|ccccc|cc|cccc}
    \toprule
    Method & NQ & TQA & zsRE & T-rex & HQA &  NQ-C & Corg & Drop & Squad & SWDE & FDA  \\ 
    \midrule
    \multicolumn{5}{l}{\textbf{Llama2 7B}} \\
    \midrule
    AdaCAD + \wocontext & 44.9 & 71.4 & 60.2 & 58.0 & 48.8 & 35.9 & 13.8 & 34.4 & 54.5 & 81.3 & \textbf{77.0} \\ 
    AdaCAD + \wcontext & \textbf{54.2} & \textbf{75.0} & \textbf{67.7} & \textbf{62.4} & \textbf{51.5} & \textbf{73.3} & \textbf{15.4} & \textbf{39.8} & \textbf{55.6} & \textbf{91.7}& 70.7\\
    \midrule
    CORG + \wocontext & 42.3 & 69.0 & 51.2 & \textbf{69.8} & 45.5 &  54.1 &22.0& 33.8 & 42.9  & \textbf{82.3} & 73.4\\ 
    CORG + \wcontext & \textbf{55.8} & \textbf{72.1} & \textbf{65.1} & \text{60.2} & \textbf{49.0} & \textbf{75.1} & \textbf{24.1} & \textbf{38.7} & \textbf{58.3} & 81.0 & \textbf{76.3} \\ 
    \midrule
    \midrule
    \multicolumn{5}{l}{\textbf{Llama3.1 8B}} \\
    \midrule
    AdaCAD + \wocontext & 49.7 & 67.5 & 63.6 & 62.1 & 49.7 & 70.7 & 14.2 & 39.0 & 56.8 & \textbf{94.3} & 79.9 \\ 
    AdaCAD + \wcontext & \textbf{51.7} & \textbf{70.1} & \textbf{65.8} & \textbf{63.9} & \textbf{54.3} & \textbf{84.7} & \textbf{15.9} & \textbf{44.3} & \textbf{56.9} & 93.0& \textbf{82.9} \\
    \midrule
    CORG + \wocontext & \textbf{48.0} & 70.2 & 57.5 & 62.7 & 48.0 & 60.7 & 21.4 & 39.0&64.9&92.0&74.0\\
    CORG + \wcontext &46.2 & \textbf{72.6} & \textbf{62.6} & \textbf{63.4} & \textbf{50.6} & \textbf{72.3} & \textbf{28.4} & \textbf{44.0} & \textbf{69.5} &\textbf{95.0} & \textbf{80.9} \\
    \bottomrule
    \end{tabular}

\label{table: full_method_compare}
\end{table*}

To examine the practical application and benefits of our analysis, we evaluate whether \wcontext further improves grounding performance over \wocontext when used as the LLM for inference-time grounding approaches.
We experiment over two inference-time grounding approaches: AdaCAD~\citep{wang2024adacad}, a decoding-based approach that improves grounding by adjusting the output distribution through logit weighting between parametric and contextual knowledge; and CORG~\citep{corg}, a pipeline-based framework designed for contexts involving complex, interrelated facts—settings where language models often struggle. 

Table~\ref{table: full_method_compare} presents the performance of \wocontext and \wcontext when combined with inference-time grounding techniques.
We find that using \wcontext as the LLM for such methods consistently improves performance across all settings compared to \wocontext.
For example, using \wcontext with AdaCAD yields an average absolute improvement of 5.3 points and pairing it with CORG gives a 6.5 point gain compared to using \wocontext as the LLM.

\subsection{Routing Input Based on Context Availability using LoRA}
\label{app_sub: lora}
\paragraph{LoRA-based training shows trends consistent with full-parameter training}

Table~\ref{table_app: Lora} shows the performance of models trained with LoRA~\citep{hu2022lora} using either context-free instruction tuning data (\wocontext) and context-augmented data (\wcontext), with Llama3.1 8B as the base model.
For the experiment, we trained only the LoRA parameters, using a rank of 16, an alpha of 32, and a dropout rate of 0.05. 
The trend tends to be similar with when training the full parameter; \wcontext consistently outperforms \wocontext, with a larger performance gain on NQ-C and CORG (+8.9\%) compared to the others (+2.0\%).

\begin{figure}[t!]
    \centering
    \begin{minipage}{\linewidth}
        \centering
        \includegraphics[width=\linewidth]{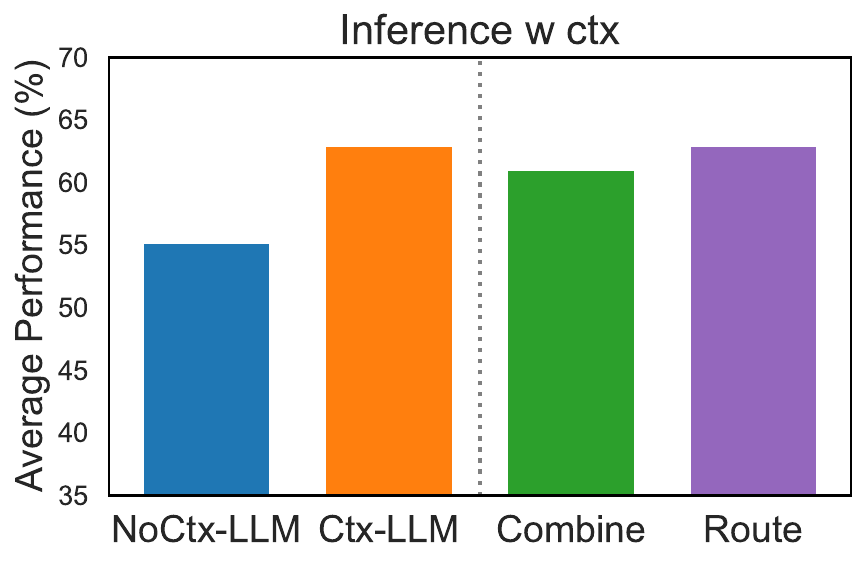}
    \end{minipage}
    \hfill
    \begin{minipage}{\linewidth}
        \centering
        \includegraphics[width=\linewidth]{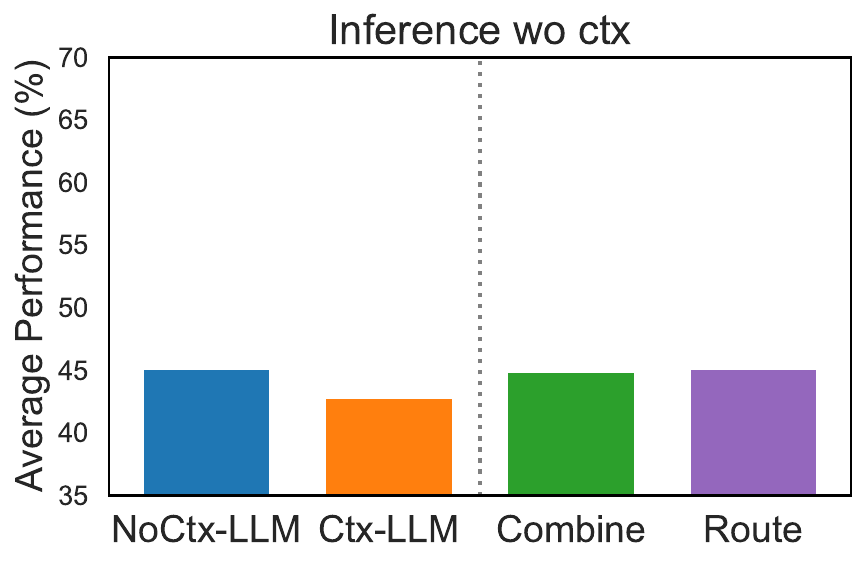}
    \end{minipage}
    \caption{
        Avg. performance across nine datasets (NQ, TQA, zsRE, T-rex, HQA, Drop, Squad, SWDE, and FDA).
        Inference \textit{w ctx} (top figure) shows performance with relevant context provided at inference time; the Inference \textit{wo ctx} (bottom figure) shows performance without relevant context.
    }
    \label{app_fig: lora_practical}
\end{figure}

\paragraph{Routing Inputs Based on Context Availability}
Figure~\ref{app_fig: lora_practical} shows a trend similar to that observed with full parameter training.
\textit{Routing} provides a strong balance, achieving robust performance in both settings, an average of 54.5, compared to 50.2 for \wocontext and 52.9 for \wcontext. 
Combining the two datasets (\textit{Combine}) also yields a good balance, with an average of 53.0.
These results suggest that when full-parameter training is computationally expensive, LoRA-based routing is an efficient alternative, especially for scaling to larger models. 

\begin{figure}[t!]
    \centering
    \begin{minipage}{\linewidth}
        \centering
        \includegraphics[width=\linewidth]{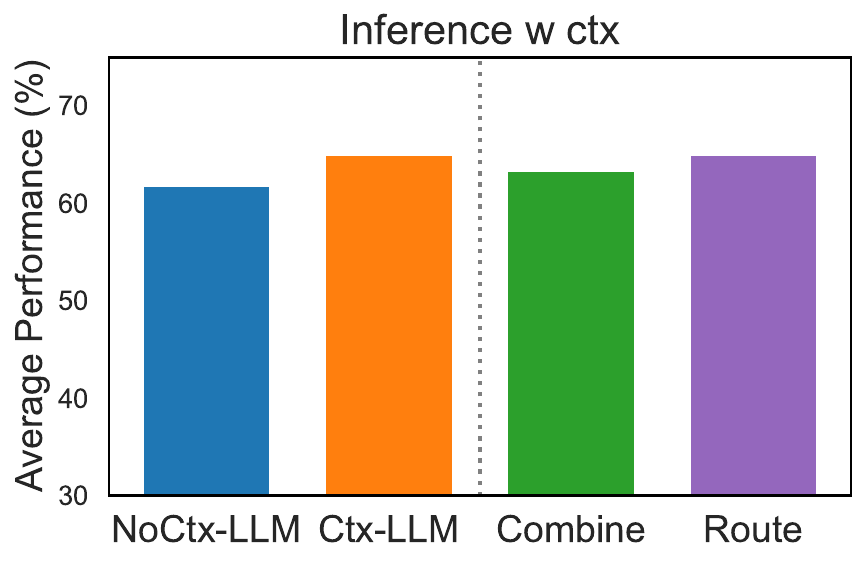}
    \end{minipage}
    \hfill
    \begin{minipage}{\linewidth}
        \centering
        \includegraphics[width=\linewidth]{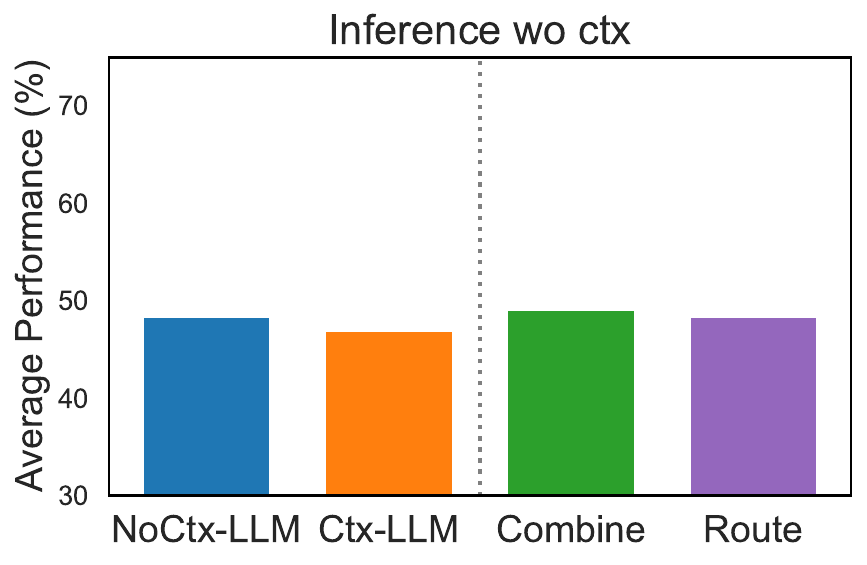}
    \end{minipage}
    \caption{
        Avg. performance across nine datasets (NQ, TQA, zsRE, T-rex, HQA, Drop, Squad, SWDE, and FDA).
        Inference \textit{w ctx} (top figure) shows performance with relevant context provided at inference time; the Inference \textit{wo ctx} (bottom figure) shows performance without relevant context. 
    }
    \label{fig: practical}
\end{figure}

\subsection{Effect of varying the ratio of context-augmented and context-free training examples}
Figure~\ref{fig: instance_ratio} shows the performance of models trained with the same total number of training examples but with different proportions of context-augmented data. 
Performance when inference is done with relevant context added (\textit{with context}) increases as the proportion of context-augmented training examples grows.
50\% mix provides the best overall balance, maintaining strong performance both with and without context.

\section{CheckList}
\subsection{Potential Risk}
A model with strong grounding ability may also reliably ground on incorrect or harmful context, potentially amplifying misinformation if the provided evidence is flawed.
However, we expect this risk can be mitigated by applying robust filtering and validation of external context before it is supplied to the model.

\subsection{LLM Usage}
We used the free version of ChatGPT-4o to assist with grammar checking during the writing of this paper.

\end{document}